\documentclass[10pt,journal,compsoc]{IEEEtran}
\usepackage{array}
\usepackage{graphicx}
\usepackage{amsmath}
\usepackage{amssymb}
\usepackage{enumerate}
\usepackage{multirow}
\usepackage{threeparttable}
\usepackage[normalem]{ulem}
\usepackage{xcolor}
\usepackage{hyperref}
\hypersetup{
    colorlinks=true,
    linkcolor=blue,
    filecolor=magenta,      
    urlcolor=blue,
}
\newcommand\eg{\emph{e.g.}} 
\newcommand\ie{\emph{i.e.}}

\newcommand\etal{\emph{et al.}}

\newcommand{\vB}{\mathbf{B}}
\newcommand{\vA}{\mathbf{A}}

\newcommand{\vL}{\mathbf{L}}

\newcommand{\vE}{\mathbf{E}}
\newcommand{\vM}{\mathbf{M}}

\newcommand{\vr}{\mathbf{r}}
\newcommand{\vw}{\mathbf{w}}
\newcommand{\vx}{\mathbf{x}}
\newcommand{\vy}{\mathbf{y}}
\newcommand{\calJ}{\mathbf{J}}

\newcommand{\calT}{\mathcal{T}}
\newcommand{\calS}{\mathcal{S}}

\def\rc#1{{\color{black}{}{{#1}}{}}}
\def\rcf#1{{\color{black}{}{{#1}}{}}}

\ifCLASSOPTIONcompsoc
  \usepackage[nocompress]{cite}
\else
  \usepackage{cite}
\fi

\ifCLASSINFOpdf
\else
\fi

\begin{document}

\title{High Frame Rate Video Reconstruction based on an Event Camera} 
\author{Liyuan Pan, Richard Hartley, Cedric Scheerlinck, Miaomiao Liu, Xin Yu, and Yuchao Dai\\ 

\IEEEcompsocitemizethanks{\IEEEcompsocthanksitem Liyuan Pan, Richard Hartley, Cedric Scheerlinck, Miaomiao Liu are with the Research School of Engineering, Australian National University, Canberra, Australia and Australian Centre for Robotic Vision. \protect\\
Xin Yu is with the Faculty of Engineering and Information Technology, University of Technology Sydney, Australia. \protect\\
Yuchao Dai is with the School of Electronics and Information, Northwestern Polytechnical University, Xi'an, China. \protect\\
E-mail: \tt\small{\{liyuan.pan\}}@anu.edu.au.
}
}

\markboth{Journal of \LaTeX\ Class Files,~Vol.~14, No.~8, November~2020}%
{Shell \MakeLowercase{\textit{et al.}}: Bare Demo of IEEEtran.cls for Computer Society Journals}

\IEEEtitleabstractindextext{%

\begin{abstract}
Event-based cameras measure intensity changes (called `{\it events}') with microsecond accuracy under high-speed motion and challenging lighting conditions.
With the `active pixel sensor' (APS), \rcf{the `Dynamic and Active-pixel Vision Sensor' (DAVIS)} allows the simultaneous output of intensity frames and events.
However, the output images are captured at a relatively low frame rate and often suffer from motion blur.
A blurred image can be regarded as the integral of a sequence of latent images, while events indicate changes between the latent images.
Thus, we are able to model the blur-generation process by associating event data to a latent sharp image.
Based on the abundant event data alongside a low frame rate, easily blurred images, we propose a simple yet effective approach to reconstruct high-quality and high frame rate sharp videos.
Starting with a single blurred frame and its event data \rc{from DAVIS}, we propose the \textbf{Event-based Double Integral (EDI)} model and solve it by adding regularization terms.
Then, we extend it to \textbf{multiple Event-based Double Integral (mEDI)} model to get more smooth results based on multiple images and their events. Furthermore, we provide a new and more efficient solver to minimize the proposed energy model.
By optimizing the energy function, we achieve significant improvements in removing blur and the reconstruction of a high temporal resolution video.
The video generation is based on solving a simple non-convex optimization problem in a single scalar variable.
Experimental results on both synthetic and real datasets demonstrate the superiority of our \textbf{mEDI} model and optimization method compared to the state-of-the-art.
\end{abstract}
\begin{IEEEkeywords}
Event Camera (\rc{DAVIS}), Motion Blur, High Temporal Resolution Reconstruction, mEDI Model, Fibonacci Sequence.
\end{IEEEkeywords}
}

\maketitle
\IEEEdisplaynontitleabstractindextext
\IEEEpeerreviewmaketitle

\IEEEraisesectionheading{\section{Introduction}\label{sec:introduction}}

\begin{figure*}[ht]
\begin{center}
\resizebox{1.01\textwidth}{!}{
\begin{tabular}{cccc}
\includegraphics[width=0.245\textwidth]{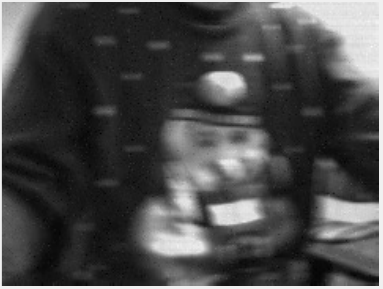}
&\includegraphics[width=0.245\textwidth]{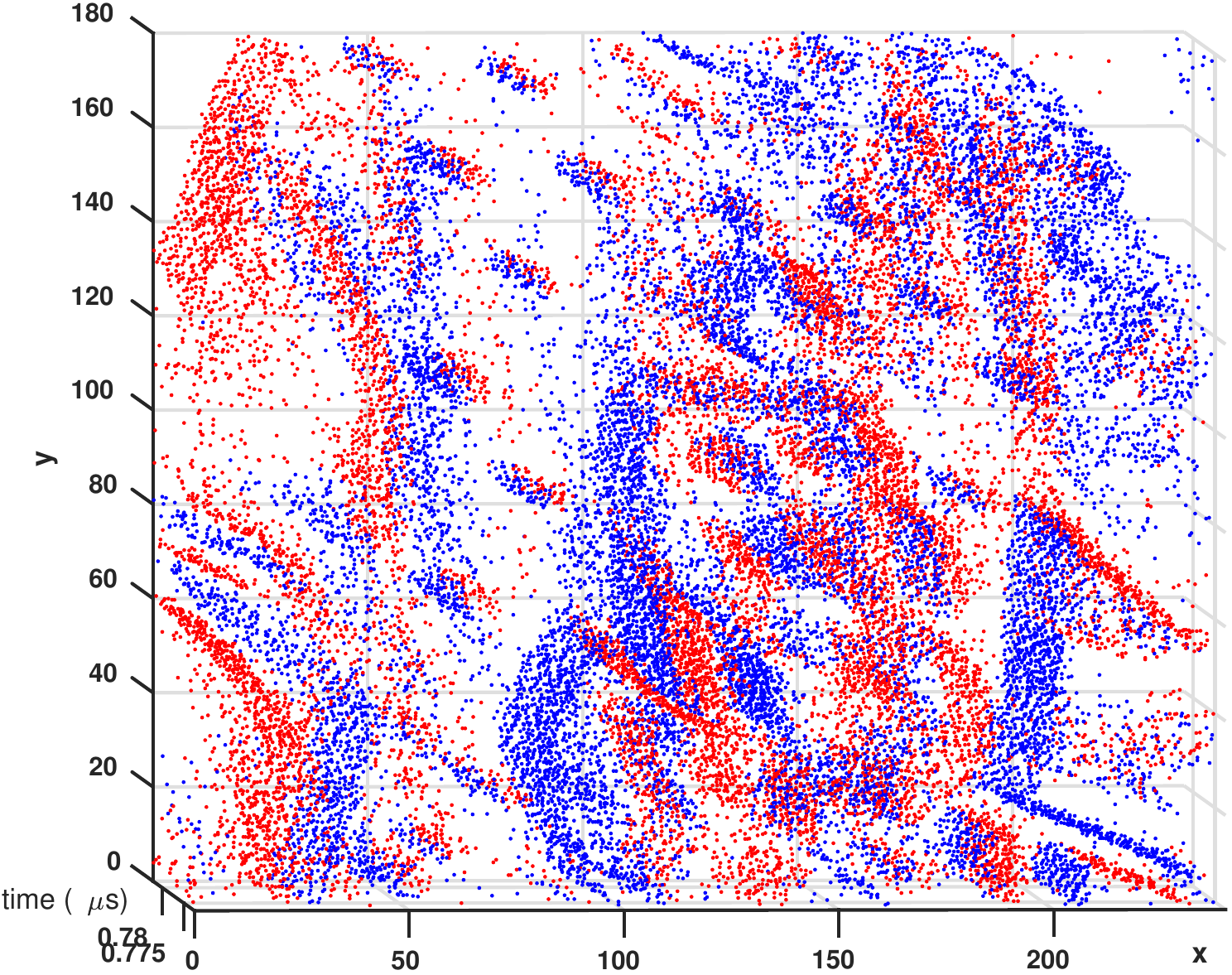}
&\includegraphics[width=0.245\textwidth]{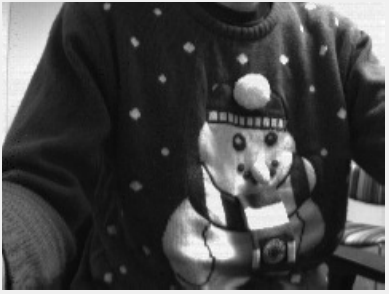}
&\includegraphics[width=0.245\textwidth]{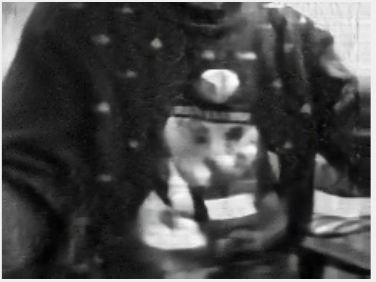}\\[0.05in]
(a) The Blurred Image  
&(b) The Events
&(c) Another View of the Sweater 
&(d) Tao \etal \cite{Tao_2018_CVPR} \\[0.05in]
\includegraphics[width=0.245\textwidth]{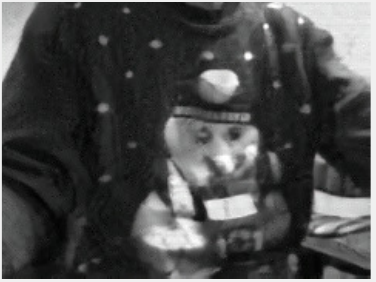}
&\includegraphics[width=0.245\textwidth]{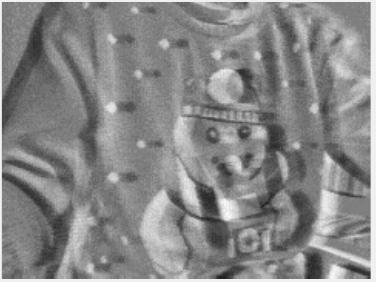}
&\includegraphics[width=0.245\textwidth]{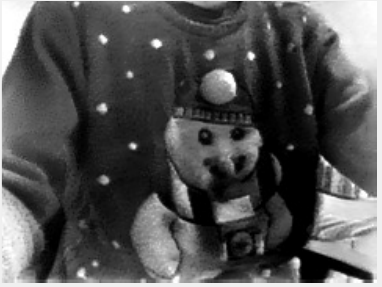}
&\includegraphics[width=0.245\textwidth]{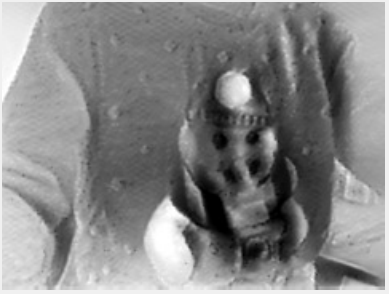}
\\
(e) Jin \etal \cite{Jin_2018_CVPR}
&(f) {\small\begin{tabular}[c]{@{}c@{}}Scheerlinck \etal \cite{Scheerlinck18accv}\\ (events only)\end{tabular}}
&(g) E2VID ($\times 10$) \cite{Rebecq_2019_CVPR}
&(h) E2VID ($\times 100$) \cite{Rebecq_2019_CVPR}\\
\includegraphics[width=0.245\textwidth]{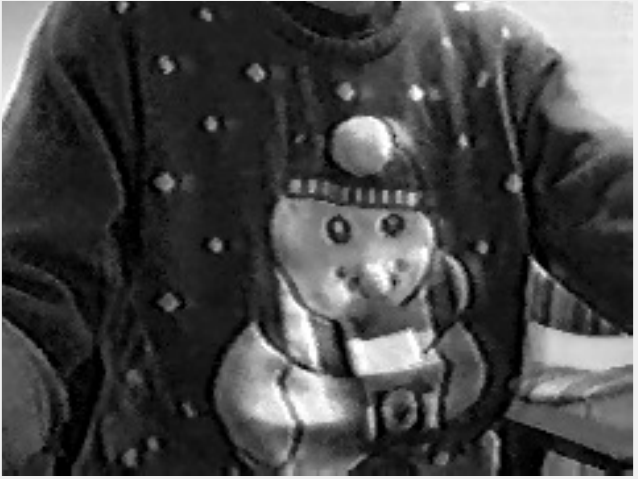}
&\includegraphics[width=0.245\textwidth]{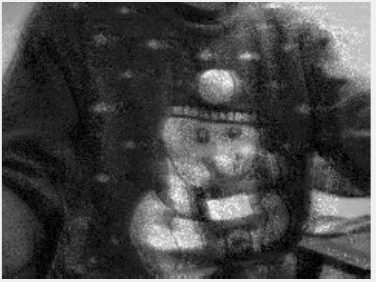}
&\includegraphics[width=0.245\textwidth]{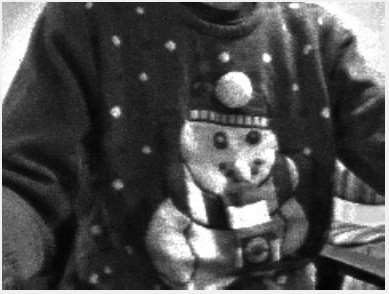}
&\includegraphics[width=0.245\textwidth]{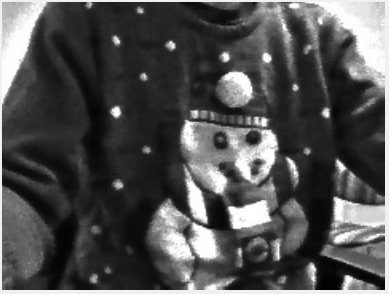}\\[0.05in]
(i) Our EDI ($\times 100$) \cite{Pan_2019_CVPR}
& (j) Scheerlinck \etal \cite{Scheerlinck18accv} 
&(k) Our mEDI ($\times 10$) 
&(l) Our mEDI ($\times 100$)\\
\end{tabular}
}
\end{center}
\vspace{-3mm}
 \caption{\em \label{fig:eventdeblur}
\rc{Deblurring and reconstruction results of our method compared with the state-of-the-art methods on our real {\it blur event dataset}. 
(a) The input blurred image. 
(b) The corresponding event data. 
(c) A sharp image for the sweater captured as a reference for colour and shape (a real blurred image can hardly have its ground truth sharp image).
(d) Deblurring result of Tao \etal \cite{Tao_2018_CVPR}. 
(e) Deblurring result of Jin \etal \cite{Jin_2018_CVPR}. Jin uses video as training data to train a supervised model to perform deblur, where the video can also be considered as similar information as the event data. 
(f) Reconstruction results of Scheerlinck \etal \cite{Scheerlinck18accv} from only events. 
(g) Reconstruction results of Rebecq \etal \cite{Rebecq_2019_CVPR} from only events. Based on their default settings, the time resolution of the reconstructed video is around $\times 10$ times higher than the time resolution of the original video. 
(h) Reconstruction results of Rebecq \etal \cite{Rebecq_2019_CVPR} from only events. The time resolution here is around $\times 100$.
(i) Reconstruction result of Pan \etal \cite{Pan_2019_CVPR} from combining events and a single blurred frame.
(j)Reconstruction results of Scheerlinck \etal \cite{Scheerlinck18accv} from events and images. 
(k)-(l) Our reconstruction result from combining events and multiple blurred frame at different time resolution. Our result preserves more abundant and faithful texture and the consistency of the natural image. (Best viewed on screen).}
}
\end{figure*}

\IEEEPARstart{E}{vent} cameras (such as the Dynamic Vision Sensor (DVS) \cite{lichtsteiner2008128} and the DAVIS \cite{brandli2014240}) are sensors that asynchronously measure intensity changes at each pixel independently with microsecond temporal resolution (if nothing moves in the scene, no events are triggered).
The event stream encodes the motion information by measuring the precise pixel-by-pixel intensity changes. Event cameras are more robust to low lighting and highly dynamic scenes than traditional cameras since they are not affected by under/over exposure associated with a synchronous shutter. 

Due to the inherent differences between event cameras and standard cameras, existing computer vision algorithms designed for standard cameras cannot be applied to event cameras directly.
Although the DAVIS \cite{brandli2014240} can provide simultaneous output of intensity frames and events, there still exist major limitations with current DAVIS cameras:
\vspace{-2mm}
\begin{itemize}
\item {\bf Low frame rate intensity images:}
In contrast to the high temporal resolution of event data ($\geq 3\mu$s frame rate), the current \rc{DAVIS} only output low frame rate intensity images ($\geq \rc{20}m$s temporal resolution).
    
\vspace{0.5mm}

\item {\bf Inherent blur effects:} When recording highly dynamic scenes, motion blur is a common issue due to the relative motion between the camera and the scene. The output of the intensity image from the APS tends to be blurry.  
\end{itemize}

To address these challenges, various methods have been proposed by reconstructing high frame rate videos. Existing methods can be in general categorized as: 
\begin{enumerate}[1)]

\item Event-only solutions \cite{Bardow16cvpr, Rebecq_2019_CVPR, Wang_2019_CVPR, Scheerlinck19ral}, where the results tend to lack the texture and consistency of natural videos \rcf{(especially for scenes with a static background or a slowly moving background/foreground )}, as they fail to use the complementary information contained in low frame rate intensity images;
\vspace{+0.5mm}
\item Events and intensity images combined solutions \cite{Scheerlinck18accv, shedligeri2019photorealistic, Brandli14iscas}, which build upon the interaction between both sources of information.  
However, these methods fail to address the blur issue associated with the captured image frame. Therefore, the reconstructed high frame rate videos can be degraded by blur.
    
\end{enumerate}

Contrary to existing \rcf{`image + event' based} methods that ignore the blur effect in the image,  
or discard it entirely, we give an alternative insight into the problem. While blurred frames cause undesired image degradation, they inherently encode the relative motion between the camera and the observed scene, and the integral of multiple images during the exposure time. Taking full advantage of the encoded information in the blurred image would benefit the reconstruction of high frame rate videos.

To tackle above problems, in our previous work \cite{Pan_2019_CVPR}, we propose an {\textbf{Event-based Double Integral (EDI)}} model to fuse an image (even with blur) with its event sequence to reconstruct a high frame rate, blur-free video.
Our {\bf EDI} model naturally relates the desired high frame rate sharp video, the captured intensity frame and event data.
Based on the {\bf EDI} model, high frame rate video generation is as simple as solving a non-convex optimization problem in a single scalar variable.

As the {\bf EDI} model is based on a single image, noise from the event data can easily degrade the quality of reconstructed videos, especially at transitions between images.
To mitigate accumulated noise from events, we limit the integration to a small time interval around the centre of the exposure time, allowing us to reconstruct a small video segment associated with one image. The final video is obtained by stitching all the video segments together. However, this still result in flickering, especially when the camera and objects have larger relative motion. 
\rcf{In addition, the regularization terms (with extra weight parameters) are included in the energy function when solving the contrast threshold for our EDI model. 
Thus, we extended our \textbf{EDI} model to a \textbf{multiple Event-based Double Integral (mEDI)} one to handle discontinuities at the boundaries of reconstructed video segments and develop a simple yet effective optimization solution.} Later in our experiments, it shows the significant improvement in the smoothness and quality of the reconstructed videos.

In this paper, we first introduce our previous approach (the EDI model) in Sec.~\ref{sec:EDI}. Then, we build an extension framework based on multiple images and describe the approach in Sec.~\ref{sec:mEDI}. \rc{Jointly optimizing \emph{multi-frames for generating long video sequences} significantly alleviates the flickering problem for the generated videos, whereas EDI treats each image individually and may suffer flicking artifacts.}

\vspace{+1mm}
The extensions are as follows:
\vspace{-0.5mm}

\begin{enumerate}[1)] 
\item We propose a \textbf{multiple Event-based Double Integral (mEDI)} model to restore better high frame rate sharp videos. The model is based on multiple images (even blurred) and their corresponding events.
\vspace{+1mm}

\item Our \textbf{mEDI} is able to generate a sharp video under various types of blur by solving a single variable non-convex optimization problem, especially in low lighting condition and complex dynamic scene. 
\vspace{+1mm}

\item We develop a simple yet effective optimization solution. In doing so, we signiﬁcantly reduce the computational complexity with the Fibonacci sequence. 
\vspace{+1mm}

\item The frame rate of our reconstructed video can theoretically be as high as the event rate (200 times greater than the original frame rate in our experiment). With multiple images, the reconstructed videos preserve more abundant texture and the consistency of natural images.
\end{enumerate}

\section{Related Work}

Event cameras such as the DAVIS and DVS \cite{brandli2014240,lichtsteiner2008128} report log intensity changes, inspired by human vision. The result is a continuous, asynchronous stream of events that encodes non-redundant information about local brightness change.
Estimating intensity images from events is important. The reconstructed images grant computer vision researchers a readily available high temporal resolution, high-dynamic-range imaging platform that can be used for tasks such as face-detection \cite{Barua16wcav}, \rc{moving object segmentation~\cite{stoffregen2019event}, SLAM \cite{Cook11ijcnn,Kim14bmvc,Kim16eccv,Rebecq17ral,vidal2018ultimate}, localization~\cite{Liu_2017_ICCV,liu2019stochastic} and optical flow estimation \cite{Zhu-RSS-18,Gehrig_2019_ICCV,Stoffregen20eccv,pan2020single}}.
Although several works try to explore the advantages of the high temporal resolution provided by event cameras \cite{zhou2018semi,Zhu17cvpr,Gehrig18eccv,Kueng16iros,gallego2019event,brandli2014real}, how to make the best use of the event camera has not yet been fully investigated. \rc{In this section, we review image reconstruction from event-based methods, and images and event combined methods. We further discuss works on image deblurring.}

\vspace{2mm}
{\noindent{\bf Event-based image reconstruction.}} A typical way is done by processing a spatio-temporal window of events. Taking a spatio-temporal window of events imposes a latency cost at minimum equal to the length of the time window, and choosing a time-interval (or event batch size) that works robustly for all types of scenes is not trivial. Barua \etal \cite{Barua16wcav} generate image gradients by dictionary learning and obtain a logarithmic intensity image via Poisson reconstruction. Bardow \etal \cite{Bardow16cvpr} simultaneously  optimise  optical flow and  intensity estimates within a fixed-length, sliding spatio-temporal window using the primal-dual algorithm \cite{Posch10isscc}. 
\rcf{Cook \etal~\cite{Cook11ijcnn} integrate events into interacting maps to recover intensity, gradient, and optical flow while estimating global rotating camera motion. }
Kim \etal \cite{Kim14bmvc} reconstruct high-quality images from an event camera under a strong assumption that the only movement is pure camera rotation, and later extend their work to handle 6-degree-of-freedom motion and depth estimation \cite{Kim16eccv}.
Reinbacher \etal \cite{Reinbacher16bmvc} integrate events over time while periodically regularising the estimate on a manifold defined by the timestamps of the latest events at each pixel. 
\rc{Optimization based event-only methods (\ie~without the process of learning from training data) will generate artifacts and lack of texture when event data is sparse, because they cannot integrate sufficient information from the available sparse events. }
\rc{
Recently, learning-based approaches have improved the image reconstruction quality significantly with powerful event data representations via deep learning~\cite{Rebecq_2019_CVPR,rebecq2019high, Wang_2019_CVPR,Scheerlinck20wacv}.
Rebecq~\etal~propose E2VID \cite{Rebecq_2019_CVPR}, a fully convolutional, recurrent UNet architecture to encode events in a spatio-temporal voxel grid. 
In~\cite{rebecq2019high}, Rebecq \etal~propose a recurrent network to reconstruct videos from a stream of events and they incorporate stacked ConvLSTM gates, which prevent vanishing gradients during backpropagation for long sequences.
Wang \etal~\cite{Wang_2019_CVPR} form a 3D event volume by stacking event frame in a time interval. A reconstructed intensity frame is generated by summing events at each pixel in a smaller time interval. }

To achieve more image details in the reconstructed images, several methods trying to combine events with intensities have been proposed.
The DAVIS \cite{brandli2014240} uses a shared photo-sensor array to simultaneously output events (DVS) and intensity images (APS).
\rc{
Brandli \etal \cite{Brandli14iscas} combine images and event streams from the DAVIS camera to create inter-frame intensity estimates by dynamically estimating the contrast threshold  (temporal contrast) of each event. Each new image frame resets the intensity estimate, preventing excessive growth of integration error. However, it also discards important accumulated event information. Scheerlinck \etal \cite{Scheerlinck18accv} propose an asynchronous event-driven complementary filter to combine APS intensity images with events, and obtain continuous-time image intensities.}
Shedligeri \etal \cite{shedligeri2019photorealistic} first exploit two intensity images to estimate depth. Second, they only use events to reconstruct a pseudo-intensity sequence (using method \cite{Reinbacher16bmvc}) between the two intensity images. They, taking the pseudo-intensity sequence, they estimate the ego-motion using visual odometry. With the estimated 6-DOF pose and depth, they directly warp the intensity image to the intermediate location.
Liu \etal \cite{Liu17vc} assume a scene should have static background. Thus, their method needs an extra sharp static foreground image as input and the event data are used to align the foreground with the background.

\vspace{1mm}

{\noindent{\bf Image deblurring.}}
Recently, significant progress has been made in blind image deblurring.
Traditional deblurring methods usually make assumptions on the scenes (such as a static scene) or exploit multiple images (such as stereo, or video) to solve the deblurring problem. 
Significant progress has been made in the field of single image deblurring. Methods using gradient based regularizers, such as Gaussian scale mixture \cite{fergus2006removing}, $l_1 \backslash l_2$ norm \cite{krishnan2011blind}, edge-based patch priors \cite{sun2013edge,yu2014efficient} and ${l}_0$-norm regularizer \cite{xu2013unnatural,pan2019phase}, have been proposed. Non-gradient-based priors such as the color line based prior \cite{lai2015blur}, and the extreme channel (dark/bright channel) prior \cite{pan2017deblurring, yan2017image} have also been explored.
Since blur parameters and the latent image are difficult to be estimated from a single image, the single-image-based approaches are extended to use multiple images \cite{hyun2015generalized, sellent2016stereo, Pan_2017_CVPR, pan2018depth, pan2019tip}. 

Driven by the success of deep neural networks, Sun \etal \cite{sun2015learning} propose a convolutional neural network (CNN) to estimate locally linear blur kernels. Gong \etal \cite{gong2017motion} learn optical flow from a single blurred image through a fully-convolutional deep neural network. The blur kernel is then obtained from the estimated optical flow to restore the sharp image. 
Nah \etal~\cite{Nah_2017_CVPR} propose a multi-scale CNN that restores latent images in an end-to-end learning manner without assuming any restricted blur kernel model. Tao \etal \cite{Tao_2018_CVPR} propose a light and compact network, SRN-DeblurNet, to deblur the image.
However, deep deblurring methods generally need a large dataset to train the model and usually require sharp images provided as supervision. In practice, blurred images do not always have corresponding ground-truth sharp images.

\vspace{1mm}
{\noindent{\bf Blurred image to sharp video.}}
Recently, two deep learning based methods \cite{Jin_2018_CVPR,purohit2018bringing} propose to restore a video from a single blurred image with a fixed sequence length.
However, their reconstructed videos do not obey the 3D geometry of the scene and camera motion.
Although deep-learning based methods achieve impressive  performance in various scenarios, their success heavily depend on the consistency between the training datasets and the testing datasets, thus hinder the generalization ability for real-world applications. 
 
\begin{figure*}[t]
\begin{center}
\resizebox{0.955\textwidth}{!}{
\begin{tabular}{cccc}
\hspace{-0.30 cm }
\includegraphics[width=0.232\textwidth]{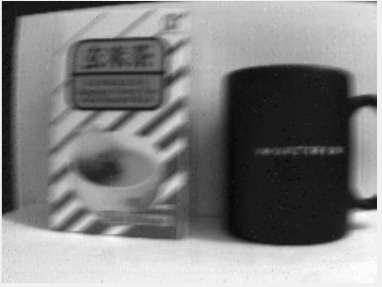}
&\includegraphics[width=0.232\textwidth]{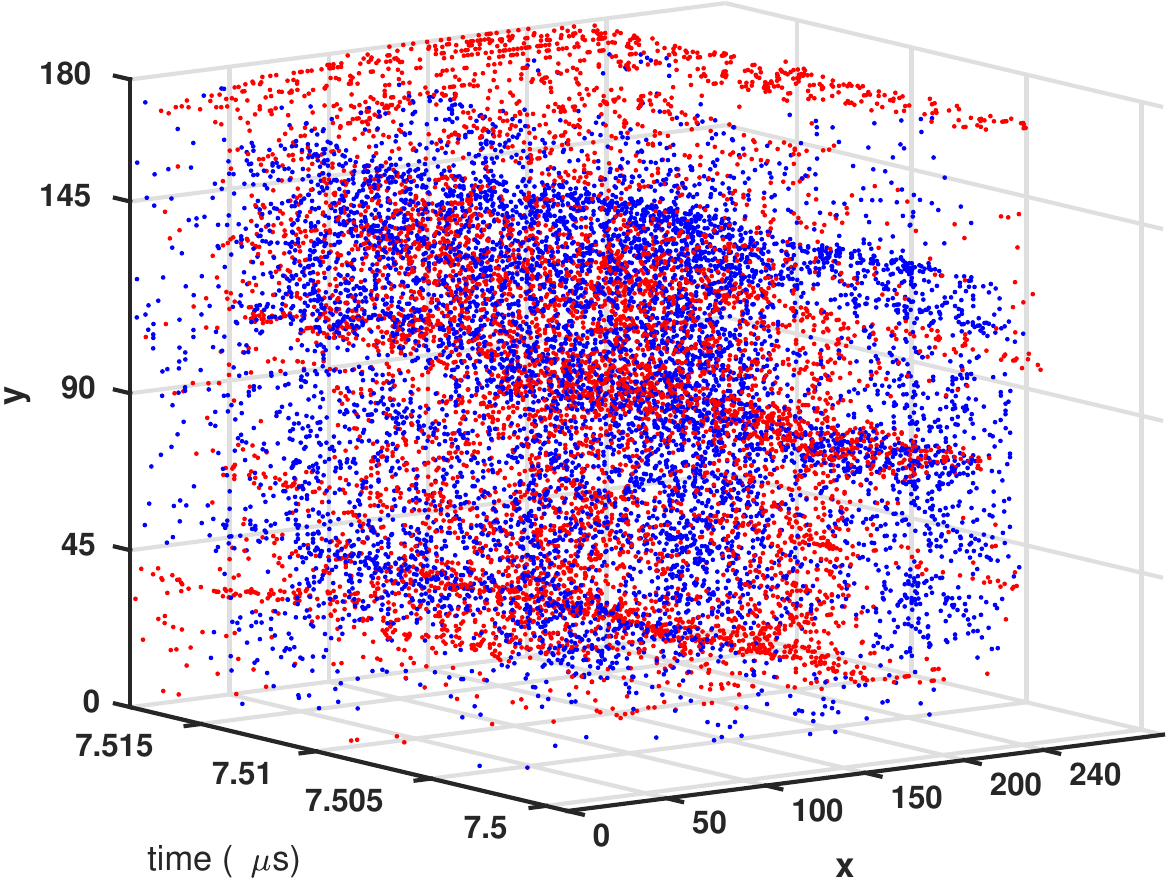}
&\includegraphics[width=0.232\textwidth]{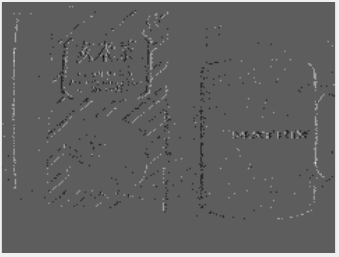}
&\includegraphics[width=0.232\textwidth]{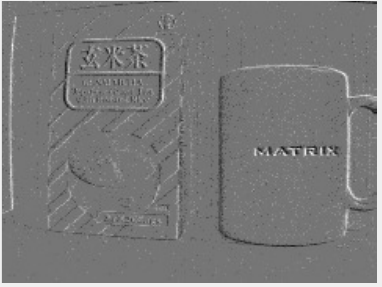}\\[0.05in]
\hspace{-0.30 cm }
(a) The Blurred Image  
&(b) The Events
&(c) $E(t) = \int e(t) \, dt$  
&(d) $\frac{1}{T}\int \exp(c\,\vE(t)) dt$\\[0.05in]
\hspace{-0.30 cm }
\includegraphics[width=0.232\textwidth]{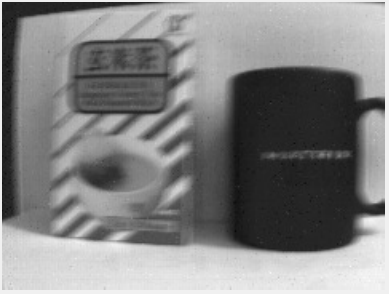}
&\includegraphics[width=0.232\textwidth]{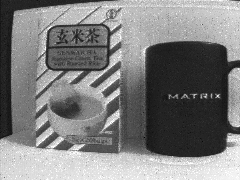}
&\includegraphics[width=0.232\textwidth]{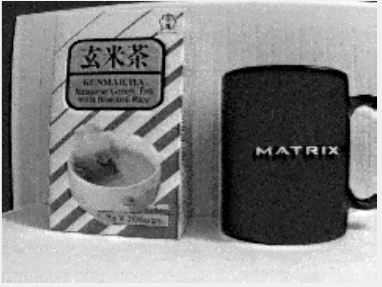}
&\includegraphics[width=0.232\textwidth]{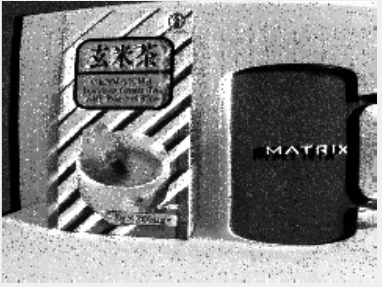}\\[0.05in]
(e) $c$ = 0.10  
&(f) $c$ = 0.22
&(g) $c$ = 0.23  
&(h) $c$ = 0.60\\
\end{tabular}
}
\end{center}
\vspace{-3mm}
\caption{\label{fig:eventsample} \em The event data and our reconstructed result, where (a) and (b) are the input of our method. (a) The intensity image from the \rc{DAVIS}.
(b) Events from the event camera plotted in 3D space-time $(x, y, t)$ (blue: positive event; red: negative event).
(c) The first integral of several events during a small time interval. 
(d) The second integral of events during the exposure time. 
(e)-(h) Samples of reconstructed image with different $c$. The value is from low (0.10), to proper (around 0.23) and high (0.60). Note, $c$ = 0.23 in (g) is the chosen automatically by our optimization process. 
}
\end{figure*}
\section{Formulation} \label{sec:EDI}

Our goal is to reconstruct a high frame rate, sharp video from a single or multiple (blurred) images and their corresponding events. In this section, we first introduce our {\bf EDI} model. 
Then, we extend it to the {\bf mEDI} model that includes multiple blurred images. Our models, both {\bf EDI} and {\bf mEDI}, can tackle various blur types and work stably in highly dynamic scenarios and low lighting conditions. 

\subsection{Event Camera Model}

Event cameras are bio-inspired sensors that asynchronously report logarithmic intensity changes \cite{brandli2014240,lichtsteiner2008128}. Unlike conventional cameras that produce full images at a fixed frame rate, event cameras trigger events whenever the change in intensity at a given pixel exceeds a preset threshold. Event cameras do not suffer from limited dynamic ranges typical of sensors with the synchronous exposure time, and capture the high-speed motion with microsecond accuracy.

Inherent in the theory of event cameras is the concept of the latent image $\vL_{xy}(t)$, denoting the instantaneous intensity at pixel $(x, y)$ at time $t$, 
related to the rate of photon arrival at that pixel. 
The latent image $\vL_{xy}(t)$ is not directly output by the camera.
Instead, the camera outputs a sequence of {\em events}, denoted by $(x,y,t,\sigma)$. 
Here, $(x, y)$ denote image coordinates, $t$ denotes the time the event takes place, and polarity $\sigma = \pm 1 $ denotes the direction (increase or decrease) of the intensity change at that pixel and time. Polarity is given by, 
\begin{align} \label{eq:log}
\sigma = \calT\left( \log \Big( \frac{\vL_{xy}(t)}{\vL_{xy}(t_\text{ref})} \Big) , c \right),
\end{align}
where $\calT(\cdot,\cdot)$ is a truncation function,
\begin{equation} \nonumber \label{eq:pkmodel}
\calT(d,c) =  
\begin{cases}
+1, & d \geq c,\\
-1, & d \leq -c.
\end{cases}
\end{equation}
Here, $c$ is a threshold parameter determining whether an event should be recorded or not, $\vL_{xy}(t)$ and $\vL_{xy}(t_\text{ref})$ denote the intensity of the pixel $(x,y)$ at time instances $t$ and $t_\text{ref}$, respectively. When an event is triggered, $\vL_{xy}(t_\text{ref})$ at that pixel is updated to a new intensity level. 
\rcf{
As described by \cite{lichtsteiner2008128}, the DVS only uses a global threshold $c$.  However, the contrast threshold of an event camera is not constant, but normally distributed. Several methods \cite{delbruck2020v2e,gallego2017event} assume that the positive and negative contrast thresholds (\ie, $c_+$ and $c_-$) exhibit different distribution noise.
We observed using a global threshold $c$, (\ie, $c_+ = c_-$ ) also yields satisfying video deblurring and high-frame rate reconstruction results while significantly simplifying the optimization procedure. Thus, we adopt a global $c$ in the following section.
}

\subsection{Intensity Image Formation}

In addition to event streams, event cameras can provide full-frame grey-scale intensity images, at a much lower rate than the event sequence.
Grey-scale images may suffer from motion blur due to their long exposure time. A general model of the blurred image formation is given by,

\begin{equation} \label{eq:avgmodel}
\vB =  \frac{1}{T}\int_{f-T/2}^{f+T/2} \vL(t) \, dt,
\end{equation}
where $\vB$ is the blurred image, equal to the average of latent images during the exposure time $[f-T/2, f+T/2]$. Let $\vL(f)$ be the snapshot of the image intensity at time $f$, the latent sharp image at the centre
of the exposure period.

\begin{figure*}[ht]
\begin{center}
\resizebox{0.96\textwidth}{!}{
\begin{tabular}{cccc}
\hspace{-0.30 cm}
\includegraphics[width=0.219\textwidth]{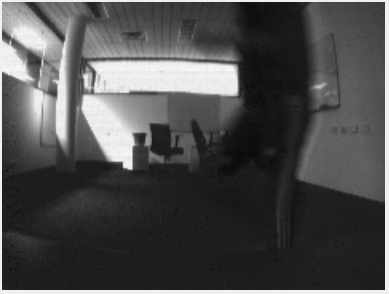}
&\includegraphics[width=0.219\textwidth]{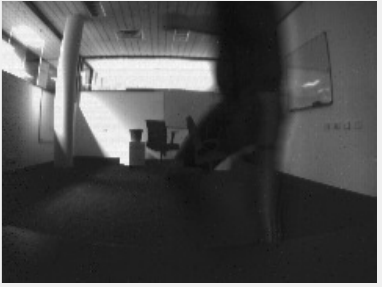}
&\includegraphics[width=0.219\textwidth]{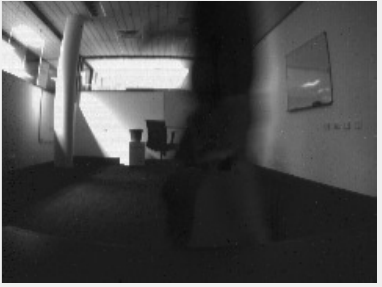}
&\includegraphics[width=0.219\textwidth]{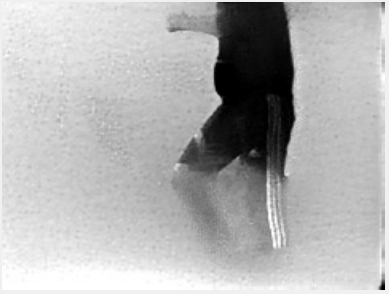}\\
\hspace{-0.30 cm }
\includegraphics[width=0.219\textwidth]{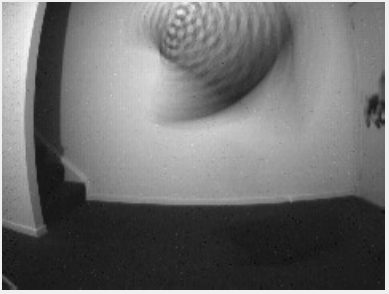}
&\includegraphics[width=0.219\textwidth]{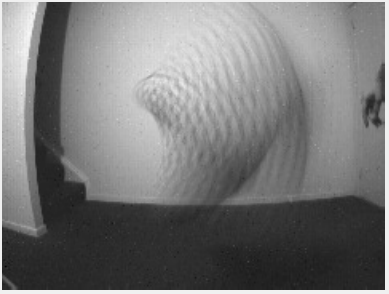}
&\includegraphics[width=0.219\textwidth]{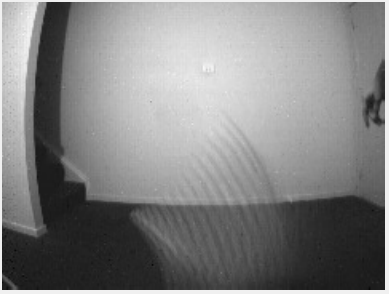}
&\includegraphics[width=0.219\textwidth]{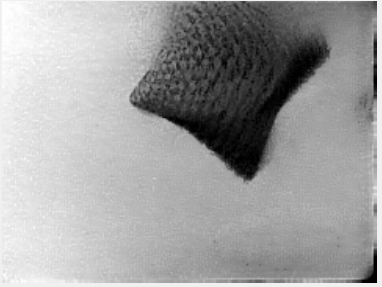}
\\
\multicolumn{3}{c}{(a) The blurred Images $\vB_0$, $\vB_1$ and $\vB_2$ (from left to right)}
& (c) Reconstructed $\vL_1$~\cite{Rebecq_2019_CVPR}\\
\hspace{-0.30 cm}
\includegraphics[width=0.219\textwidth]{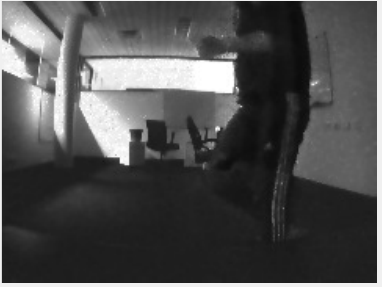}
&\includegraphics[width=0.219\textwidth]{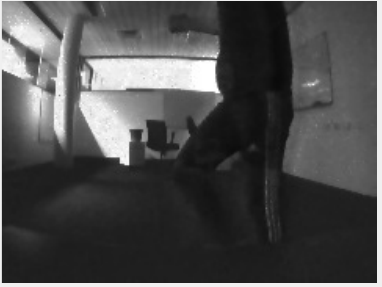}
&\includegraphics[width=0.219\textwidth]{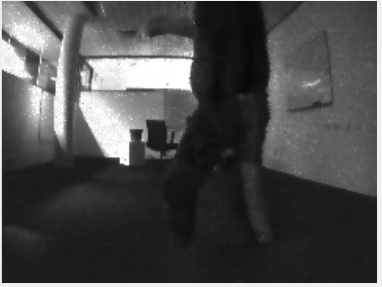}
&\includegraphics[width=0.219\textwidth]{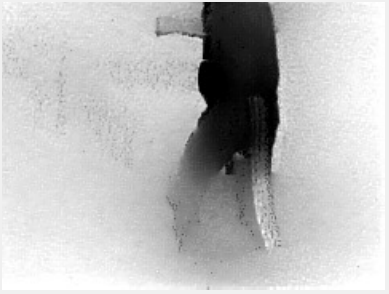}\\
\hspace{-0.30 cm}
\includegraphics[width=0.219\textwidth]{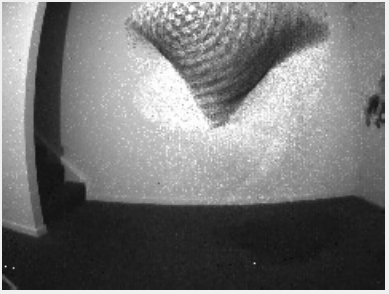}
&\includegraphics[width=0.219\textwidth]{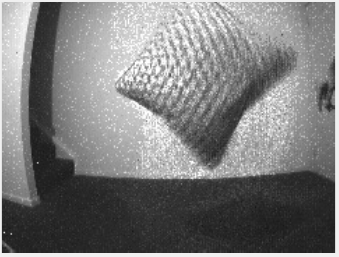}
&\includegraphics[width=0.219\textwidth]{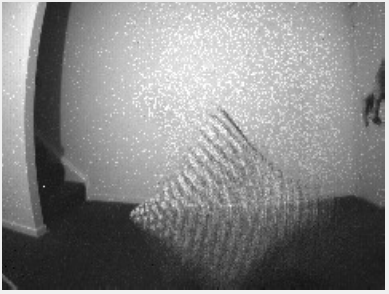}
&\includegraphics[width=0.219\textwidth]{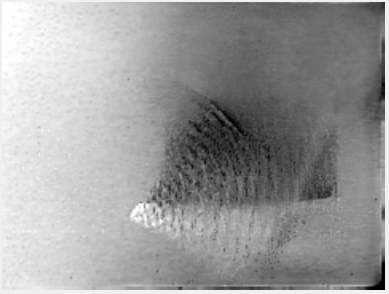}\\
\multicolumn{3}{c}{(b) Our Reconstructed Images $\vL_0$, $\vL_1$ and $\vL_2$ $(\times 100)$ (from left to right)}
&(d) Reconstructed $\vL_2$~\cite{Rebecq_2019_CVPR}\\
\end{tabular}
}
\end{center}
\vspace{-3mm}
\caption{\label{fig:multiimage} \em The examples of our reconstructed results are based on our real event dataset. The threshold $c$ is estimated automatically from three blurred images and their events based on our mEDI model. 
\rc{(a), (b) Blur image and our reconstructed Images $\vL_0$, $\vL_1$ and $\vL_2$
(c), (d) Reconstruction results of $\vL_1$ and $\vL_2$ by Rebecq \etal \cite{Rebecq_2019_CVPR} from only events. The time resolution here is around $\times 6$ based on their default settings. The time resolution of the reconstructed video by E2VID \cite{Rebecq_2019_CVPR} is around $\times 8$ to $15$ times higher than the time resolution of the original video. (Best viewed on screen).}
}
\end{figure*}

\begin{figure*}[t]
\begin{center}
\begin{tabular}{cccc}
\hspace{-0.2 cm}
\includegraphics[width=0.216\textwidth]{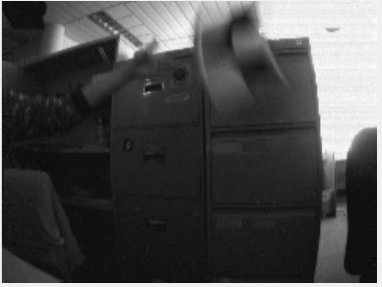}
\hspace{-0.2 cm}
&\includegraphics[width=0.216\textwidth]{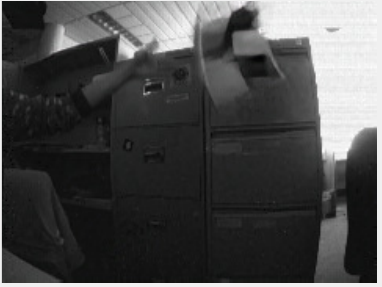}
\hspace{-0.2 cm}
&\includegraphics[width=0.216\textwidth]{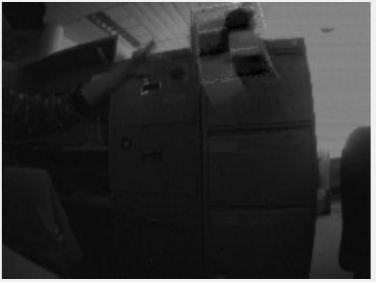}
\hspace{-0.2 cm}
&\includegraphics[width=0.216\textwidth]{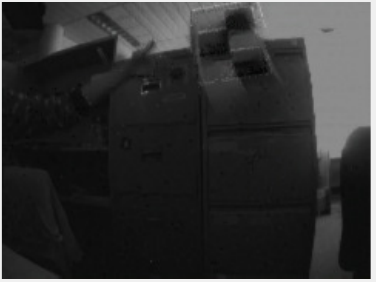}\\
\hspace{-0.2 cm}
(a) The Blurred Image 
&(b) Jin \etal \cite{Jin_2018_CVPR}  
&(c) Baseline 1
&(d) Baseline 2\\
\hspace{-0.2 cm}
\includegraphics[width=0.216\textwidth]{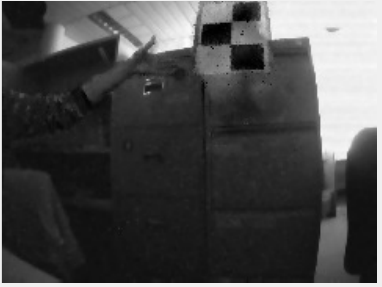}
\hspace{-0.2 cm}
&\includegraphics[width=0.216\textwidth]{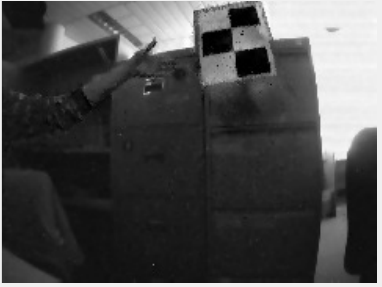}
\hspace{-0.2 cm}
&\includegraphics[width=0.216\textwidth]{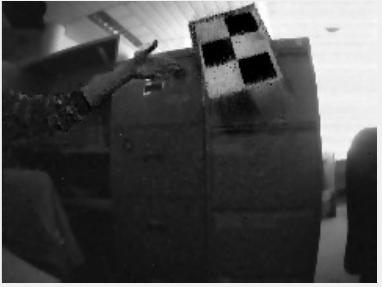}
\hspace{-0.2 cm}
&\includegraphics[width=0.216\textwidth]{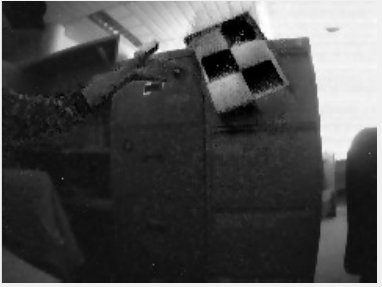}\\
\multicolumn{4}{c}{(f) Samples of Our Reconstructed Video}
\end{tabular}
\end{center}
\vspace{-3mm}
\caption{\em \label{fig:baseline} Deblurring and reconstruction results on our real {\it blur event dataset}. 
(a) Input blurred images. 
(b) Deblurring result of \cite{Jin_2018_CVPR}. 
(c) Baseline 1 for our method. We first use the state-of-the-art video-based deblurring method \cite{Jin_2018_CVPR} to recover a sharp image. Then use the sharp image as input to a state-of-the-art reconstruction method \cite{Scheerlinck18accv} to get the intensity image.
(d) Baseline 2 for our method. We first use method \cite{Scheerlinck18accv} to reconstruct an intensity image. Then use a deblurring method \cite{Jin_2018_CVPR} to recover a sharp image.
(e) Samples from our reconstructed video from $\vL(0)$ to $\vL(150)$.
}
\end{figure*}

\subsection{Event-based Double Integral Model}
We aim to recover the latent sharp intensity video by exploiting both the blur model and the event model. 
We define $e_{xy}(t)$ as a function of continuous time $t$ such that,
\begin{equation} \nonumber \label{eq:et}
e_{xy}(t) = \sigma \,\, \delta_{t_0}(t),
\end{equation}
whenever there is an event $(x, y, t_0, \sigma)$.
Here, $\delta_{t_0}(t)$ is an impulse function, with unit integral, at time $t_0$, and the sequence of events is turned into a continuous time signal, consisting of a sequence of impulses.
There is such a function $e_{xy}(t)$ for every point $(x,y)$ in the image.
Since each pixel can be treated separately, we omit the subscripts $x, y$.

\rcf{Given a reference timestamp $f$, we define $\vE(t)$ as the sum of events between time $f$ and $t$},
\begin{equation} \nonumber \label{eq:Etwithet} 
\begin{split}
\vE(t) &= \int_{f}^t e(s) ds ,\\
\end{split}
\end{equation}
which represents the proportional change in intensity between time $f$ and $t$.
Except under extreme conditions, such as glare and no-light conditions,
the latent image sequence $\vL(t)$ is expressed as,
\begin{equation}\label{eq:Ltsigma} \nonumber
\begin{split}
\vL(t)\ & = \vL(f) \, \exp( c\,  \vE(t))~. \\
\end{split}
\end{equation}
 
In particular, an event $(x,y,t,\sigma)$ is triggered when the intensity of a pixel $(x,y)$ increases or decreases by an amount $c$ at time $t$.
With a high enough temporal resolution, the intensity changes of each pixel can be segmented to consecutive event streams with different amounts of events.
We put a tilde on top of things to denote logarithm, \eg $\widetilde{\vL}(t)=\log({\vL}(t))$. Thus, we have,
\begin{equation}\label{eq:logLtwithEt}
\begin{split}
\widetilde{\vL}(t)\ & = \widetilde{\vL}(f) + c \, \vE(t).\\
\end{split}
\end{equation}

Given a sharp frame, we can reconstruct a high frame rate video from the sharp starting point $\vL(f)$ by using Eq.~\eqref{eq:logLtwithEt}. When an input image is blurred, a trivial solution would be to first deblur the image with an existing deblurring method and then to reconstruct a video using Eq.~\eqref{eq:logLtwithEt} (see Fig.~\ref{fig:baseline} for details). 
However, in this way, the event data between intensity images are not fully exploited, thus resulting in inferior performance. Moreover, none of existing deblurring methods can be guaranteed to work stably in a complex dynamic scenery. 
Instead, we propose to reconstruct the video by exploiting the inherent connection between events and blur, and present the following model.

As for the blurred image,
\begin{equation}\label{eq:blurevent} 
\begin{split}
\vB 
& = \frac{1}{T}\int_{f-T/2}^{f+T/2}  \vL(f) \,\exp \Big (c \,\vE(t)\Big)\ dt\\
& = \frac{\vL(f)}{T}\int_{f-T/2}^{f+T/2}  \exp \Big (c \int_{f}^{t} e(s) ds\Big)\ dt  ~.
\end{split}
\end{equation}

In this manner, we build the relation between the captured blurred image $\vB$ and the latent image $\vL(f)$ through the double integral of the event. We name Eq. \eqref{eq:blurevent} the \textbf{Event-based Double Integral (EDI)} model. 

We denote
\[
\calJ(c) =  \frac{1}{T}\int_{f-T/2}^{f+T/2} \exp(c \,\vE(t))dt. 
\]

Taking the logarithm on both sides of Eq.~\eqref{eq:blurevent} and rearranging it yields

\begin{equation}\label{eq:logEDIM}
\begin{split}
 \widetilde{\vL}(f) &= \widetilde{\vB} - \widetilde{\calJ}(c), 
\end{split}
\end{equation}
which shows a linear relationship between the blurred image, the latent image and integrated events in the log space.


\begin{figure*}
\begin{center}
\resizebox{0.98\textwidth}{!}{
\begin{tabular}{cccc}
\hspace{-0.3 cm }
\includegraphics[width=0.234\textwidth]{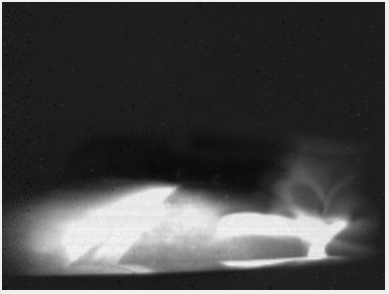}
&\includegraphics[width=0.234\textwidth]{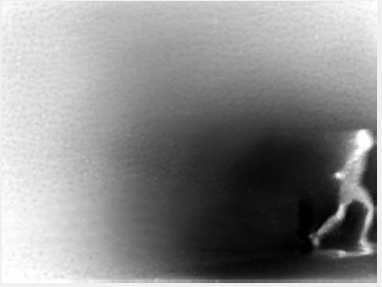}
&\includegraphics[width=0.234\textwidth]{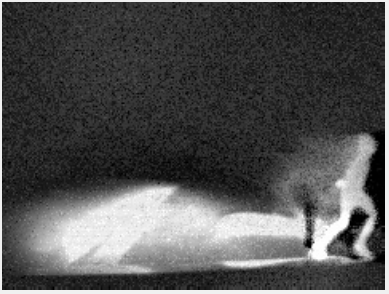}
&\includegraphics[width=0.234\textwidth]{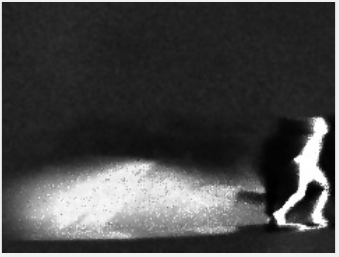}
\\[0.05in]
\hspace{-0.3 cm }
\includegraphics[width=0.234\textwidth]{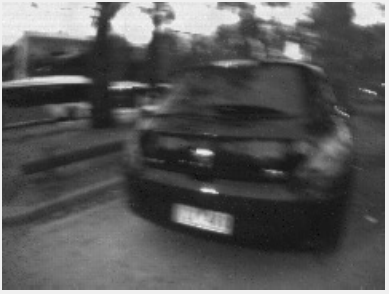}
&\includegraphics[width=0.234\textwidth]{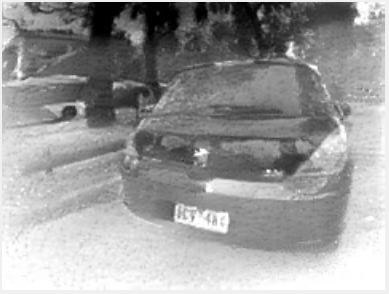}
&\includegraphics[width=0.234\textwidth]{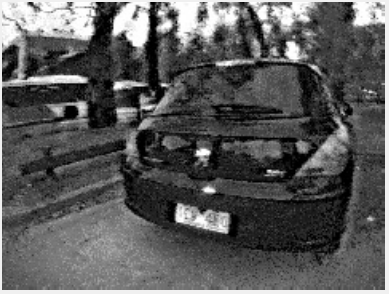}
&\includegraphics[width=0.234\textwidth]{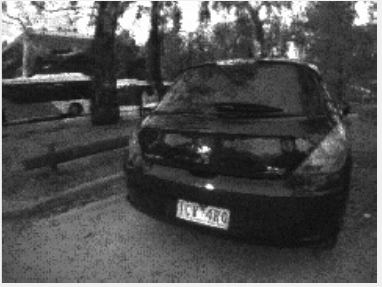}
\\
(a) The Blurred Image  
&(b) E2VID $(\times 8)$\cite{Rebecq_2019_CVPR}
&(c) Our EDI $(\times 100)$\cite{Pan_2019_CVPR}
&(d) Our mEDI $(\times 100)$
\\
\end{tabular}
}
\vspace{-3mm}
\end{center}
\caption{\label{fig:comparecvprandnow} \rc{\em Examples of reconstruction results on real event dataset. (a) The intensity image from the event camera.
(b) Reconstruction result of our E2VID \etal \cite{Rebecq_2019_CVPR} from only events. The temporal resolution is around $\times 8$ based on their default settings, while ours are $\times 100$ times higher than the original videos'.
(c) Reconstruction result of our EDI model \etal \cite{Pan_2019_CVPR} from combining events and a single blurred frame.
(d) Reconstruction result of our mEDI model from combining events and multiple blurred frames. Our method based on multiple images gets better results than our previous one based only on one single image, especially on large motion scenery and extreme light conditions.
(Best viewed on screen).}
}
\end{figure*}

\subsection{High Frame Rate Video Generation}
The right-hand side of Eq.~\eqref{eq:logEDIM} is known, apart from perhaps
the value of the contrast threshold $c$, the first term from
the grey-scale image, the second term from the event sequence, so
it is possible to compute $\widetilde\vL$, and hence $\vL$ by
exponentiation. Subsequently, from 
Eq.~\eqref{eq:logLtwithEt} the latent image $\vL(t)$ at any time may be computed.

To avoid accumulated errors of constructing a video from many frames
of a blurred video, it is more suitable to construct each frame $\vL(t)$ using the closest blurred frame.

Theoretically, we could generate a video with a frame rate as high as the DVS's event rate.  However, since each event carries little information and is subject to noise, several events must be processed together to yield a reasonable image. We generate a reconstructed frame every $50-100$ events, so for our experiment, the frame rate of the reconstructed video is usually $200$ times greater than the input low frame rate video. 
Furthermore, as indicated by Eq.~\eqref{eq:logEDIM}, the challenging blind motion deblurring problem has been reduced to a single variable optimization problem of how to find the best value of the contrast threshold $c$.

\subsection{Finding $c$ with Regularization Terms}
As indicated by Eq.~\eqref{eq:logEDIM}, the blind motion deblurring problem has been reduced to a single variable optimization problem of how to find the best value of the threshold $c$. To this end, we need to build an evaluation metric (energy function) that can evaluate the quality of the deblurred image $\vL(t)$. Specifically, we propose to exploit different prior knowledge for sharp images and the event data. 




\vspace*{2.0mm}
\noindent{\bf Edge constraint for event data. }
As mentioned before, when a proper $c$ is given, our reconstructed image $\vL(c,t)$ will contain much sharper edges compared with the original input intensity image. Furthermore, event cameras inherently yield responses at moving intensity boundaries, so edges in the latent image may be located where (and when) events occur. 
We convolve the event sequence with an exponentially decaying window, to obtain a denoised yet wide edge boundary,



\begin{equation} \nonumber \label{Mt}
\begin{split}
\vM(t) &= \int_{-T/2}^{T/2} \exp(-(|t-s|))  \, e(s) \, ds,\\
\end{split}
\end{equation}
Then, we use the Sobel filter $ \calS$ to get a sharper binary edge map, which is also applied to $\vL(c,t)$. Here, we use $\vL(c,t)$ to present the latent sharp image $\vL(t)$ with different $c$.




Here, we use cross-correlation between $ \calS(\vL(c,t))$ and $ \calS(\vM(t))$ to evaluate the sharpness of $\vL(c,t)$.
\begin{equation}
\phi_{\rm{edge}}(c) = \sum_{x, y} 
 \calS(\vL(c,t))(x, y) \cdot  \calS(\vM(t))(x, y) ~.
\end{equation}

\vspace*{2.0mm}
\noindent{\bf Intensity Image Constraint.}
Total variation is used to suppress noise in the latent image while preserving edges, and to penalize spatial fluctuations\cite{rudin1992nonlinear}. 
\begin{equation} \label{TVnorm}
\phi_{\rm{TV}}(c) = |\nabla \vL(c,t)|_{1},
\end{equation}
where $\nabla$ represents the gradient operators.

\vspace*{2.0mm}
\noindent{\bf Energy Minimization.}
The optimal $c$ can be estimate by solving Eq.~\eqref{eq:solveK1},
\begin{equation}\label{eq:solveK1}
\min_{c} \phi_{\rm{TV}}(c) + \lambda \phi_{\rm{edge}}(c),
\end{equation}
where $\lambda$ is a trade-off parameter. The response of cross-correlation reflects the matching rate of $\vL(c,t)$ and $\vM(t)$ which makes $\lambda<0$.  
This single-variable minimization problem can be solved by Golden Section Search. 





\section{Using More Than One Frame}
\label{sec:mEDI}
Though our EDI can reconstruct high frame rate videos efficiently, noise from events can easily degrade the quality of reconstructed videos with low temporal consistency.  
In addition, regularization terms in the energy function introduce unexpected weight parameters. 
Therefore, we propose a multiple images based approach to tackle the above problems with a simple yet effective optimization solution. 


\subsection{Multiple Event-based Double Integral Model}

Suppose an event camera captures a continuing sequence of events, and also blurred images, $\vB_i$ for $i = 0 ,\cdots,n$.
Assume that the exposure time is $T$ and the reference frame $\vB_i$ is at time $f_i$.
Each $\vB_i$ is associated with a latent image $\vL_i(f_i)$ and is generated as an integral of $\vL_i(t)$ over the exposure interval $[f_i-T/2, f_i+T/2]$. 
In addition, we rewrite $\vE(t)$, $\vL(t)$ and $\calJ(c)$ for the $i^{th}$ frame as
\begin{equation}  \nonumber
\begin{split}
\vE_i(t) &=  \int_{f_i}^t e(s) ds
\\
{\vL}_i(t) & = {\vL}_i(f_i) \,\exp( c \, \vE_i(t))
\\
\calJ_i(c) &=  \frac{1}{T}\int_{f_i-T/2}^{f_i+T/2} \exp(c \,\vE_i(t)) dt.
\\
\end{split}
\end{equation}

The {\bf EDI} model in Eq.~\eqref{eq:logEDIM} in section \ref{sec:EDI} gives
\begin{equation}\label{eq:mmlogEDIM}
\begin{split}
\widetilde{\vB}_i = \widetilde{\vL}_i(f_i) + \widetilde{\calJ}_i(c),\\
\end{split}
\end{equation}
for each blurred image in the sequence.
We use $\vL_i$ to represent $\vL_i(f_i)$ in the following section. Then,
Eq.~\eqref{eq:mmlogEDIM} is written as
\begin{equation}\label{eq:mmmlogEDIM}
\begin{split}
\widetilde{\vB}_i & = \widetilde{\vL}_i + \widetilde{\calJ}_i(c) = \widetilde{\vL}_i + a_i ~.
\end{split}
\end{equation}
The latent image $\vL_{i+1}$ is formed from latent image $\vL_i$ by
integrating events over the period $[f_i, f_{i+1}]$, which gives
\begin{equation}\label{eq:mmmlogEDIM2}
\begin{split}
\widetilde{\vL}_{i+1} 
& = \widetilde{\vL}_i + c \, \int_{f_i}^{f_{i+1}} e(s) ds\\
& = \widetilde{\vL}_i + b_i.\\
\end{split}
\end{equation}

This describes the {\bf mEDI} model based on multiple images and their events.
\begin{equation} \label{eq:mlogEDIM} 
\begin{split}
&\widetilde{\vL}_i = \widetilde{\vB}_i - a_i\\
\widetilde{\vL}_{i+1} - &\widetilde{\vL}_i 
 = b_i.\\
\end{split}
\end{equation}
The known values are $\widetilde{\vB}_i$, whereas the unknowns are $\widetilde{\vL}_i$, $a_i$ and $b_i$. These quantities are associated with a single pixel and we solve for each pixel independently.
We therefore obtain a set of linear equations based on Eq.~\eqref{eq:mlogEDIM} as 
\begin{equation} \label{eq:axb}
\left[
\begin{array}{ccccc}
    1  & -1 &    &    &   \\
       & 1  &-1 &    &  \\
       &     &\ddots & \ddots & \\
       &     &   &1   & -1 \\ \hline \\
    1  &     &    &    &   \\
       & 1  &    &    &  \\
       &     &1  &    & \\
       &     &   & \ddots & \\       
       &     &   &     & 1 \\ 
\end{array}
\right]
\begin{bmatrix}
\widetilde{\vL}_1  \\
\widetilde{\vL}_2  \\
\vdots \\
\widetilde{\vL}_n  \\
\end{bmatrix}
=
\begin{bmatrix}
-b_1  \\
\vdots \\
-b_{n-1}  \\
\widetilde{\vB}_1-a_1  \\
\vdots \\
\widetilde{\vB}_n-a_n  \\
\end{bmatrix},
\end{equation}
%
where $a_i$ and $b_i$ depend on $c$, but particularly $a_i$ depends on $c$ in a non-linear way.
Writing Eq.~\eqref{eq:axb} as $\vA\vx=\vw$, the least-squares solution is given by solving $\vA^{T}\vA\vx=\vA^{T}\vw$.

\subsection{$\tt{LU}$ Decomposition}
\label{sec:LU}




Because of their particular form, these equations can
be solved very efficiently as will now be shown. Expanding the 
equations $$\vA^{T}\vA\vx=\vA^{T}\vw$$ gives





\begin{equation}\label{eq:axb2}
\begin{array}{cc}
\left[
\begin{array}{ccccccc}
    2  & -1 &    &    &    &    &   \\
   -1  & 3  &  -1 &    &    &    &   \\
       &-1  & 3  &  -1 &    &    &   \\
       &    &    & \ddots  &    &    & \\
       &    &    &    &-1  & 3  &-1   \\
       &    &    &    &    &-1  &2   \\
\end{array}
\right]
\begin{bmatrix}
\widetilde{\vL}_1  \\
\widetilde{\vL}_2  \\
\vdots \\
\widetilde{\vL}_n  \\
\end{bmatrix}\\
=
\begin{bmatrix}
\widetilde{\vB}_1-a_1-b_1  \\
\widetilde{\vB}_2-a_2-b_2+b_1  \\
\vdots \\
\widetilde{\vB}_{n-1}-a_{n-1}-b_{n-1}+b_{n-2}  \\
\widetilde{\vB}_{n}-a_{n}+b_{n-1}  \\
\end{bmatrix}.
\end{array}
\end{equation}

This is a particularly easy set of equations to solve.
Since it has to be solved for each pixel, it is important to do it efficiently. 
The best way to solve Eq.~\eqref{eq:axb2} is to take the $\tt{LU}$ decomposition of the left-hand-side matrix, which has a particularly simple form. 

Let $ \vA^{T}\vw = \vr$, we writing Eq.~\eqref{eq:axb2} as $\vA^{T}\vA\vx=\vr$. The $\tt{LU}$ decomposition of $\vA^{T}\vA$ (with appropriate reordering of rows) is given by
{\small
\[
\begin{split}
&\tt{LU} = \\
&\left[
\begin{array}{ccccc}
   -2  & -5 &-13 & \cdots   &1     \\
    1  &    &    &          &0    \\
       & 1  &    &          &0     \\
       &    &\ddots     &   &\vdots   \\
       &    &    & 1        &0    \\
\end{array}
\right]
\left[
\begin{array}{ccccc}
   -1  & 3  &-1 &   &    \\
       & \ddots  & \ddots  &\ddots &   \\
       &   &-1    & 3         &-1    \\
       &    &     & -1  &2   \\
       &    &    &         &\phi_{2n-1}    \\
\end{array}
\right].\\
\end{split}
\]
}

More precisely, if the Fibonacci sequence is $1,2,3,5,8,\cdots$ and $\phi_k$ denotes the $k-$th entry of this sequence (thus $\phi(0) = 1$, $\phi(2) = 2$), then the top line of the left-hand matrix is
\[
\begin{bmatrix}
 \phi_2 &\phi_4 &\cdots &\phi_{2(n-1)} &1\\
\end{bmatrix},
\]
consisting of the even numbered entries of the Fibonacci sequence.
The entry at the bottom right of the right-hand matrix is $\phi_{2n-1}$, the next odd-numbered Fibonacci number, which is also the determinant of the original matrix.
Solving equations by $\tt{LU}$ decomposition and back-substitution is particularly simple in this case. The procedure in solving equations $\tt{LU}\vx=\vr$ is done by
solving
\[
\begin{split}
\tt{L}\vy&=\vr\\
\tt{U}\vx&=\vy.
\end{split}
\]

The solution of $\tt{L}\vy=\vr=(r_1,r_2,\cdots,r_n)^T$ is simply
\begin{equation} \nonumber
\vy = (r_2,r_3,\cdots,r_n,\sum^{n-1}_{i=1}r_i\phi_{2i})^T.
\end{equation}

The solution of $\tt{U}\vx=\vy$ is given by back-substitution from the bottom:
\begin{equation} \label{eq:x}
\begin{split}
x_n &= y_n/\phi_{2n-1}=\sum^{n-1}_{i=1}r_i\phi_{2i}/\phi_{2n-1}\\
x_{n-1} &=2x_n-r_n\\
x_{n-2} &=3x_{n-1}-x_n-r_{n-1}\\
x_{n-3} &=3x_{n-2}-x_{n-1}-r_{n-2}\\
\cdots \\
x_1 &= 3 x_2 - x_3 - r_2
\end{split}
\end{equation}

The values $x_i$ is the pixel value for latent image $\vL_i$.
If $c$ is known, then the values on the right of are dependent on $c$, and the sequence of $\vL_n$ can be computed. 

\begin{equation} \label{eq:x-r}
\begin{split}
\vL_n &= \sum^{n-1}_{i=1}r_i\phi_{2i}/\phi_{2n-1}\\
\vL_{n-1} &=2\vL_n-\widetilde{\vB}_{n}-a_{n}+b_{n-1}\\
\vL_{n-2} &=3\vL_{n-1}-\vL_n-\widetilde{\vB}_{n-1}-a_{n-1}-b_{n-1}+b_{n-2}\\
\vL_{n-3} &=3\vL_{n-2}-\vL_{n-1}-\widetilde{\vB}_{n-2}-a_{n-2}-b_{n-2}+b_{n-3}\\
\cdots \\
\vL_1 &= 3 \vL_2 - \vL_3 - \widetilde{\vB}_2-a_2-b_2+b_1
\end{split}
\end{equation}
Furthermore, the problem has been reduced to a single variable optimization problem of how to find the best value of the contrast threshold $c$. 




\begin{figure}[t]
\begin{center}
\begin{tabular}{cc}
\hspace{-0.25 cm}
\includegraphics[width=0.217\textwidth]{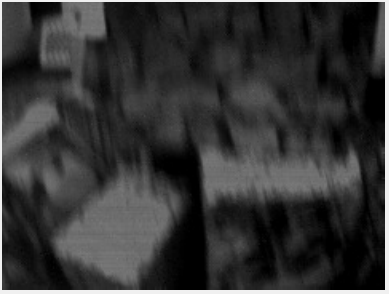}  
& \includegraphics[width=0.217\textwidth]{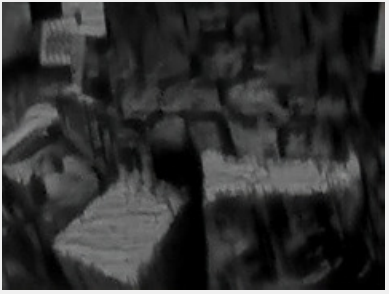}  \\
\hspace{-0.25 cm}
(a) The Blurred image
&(b) Tao \etal \cite{Tao_2018_CVPR}\\
\hspace{-0.25 cm}
\includegraphics[width=0.217\textwidth]{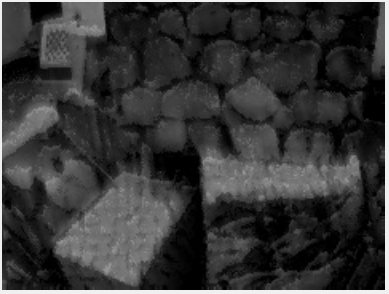}  
&\includegraphics[width=0.217\textwidth]{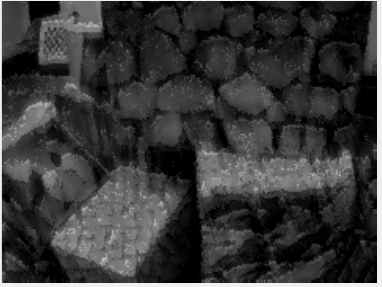} \\
\hspace{-0.25 cm}
(c) By Human Observation
&(d) By Energy Minimization\\
\hspace{-0.25 cm}
\includegraphics[width=0.217\textwidth]{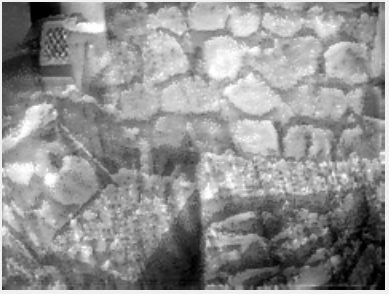}
&\includegraphics[width=0.217\textwidth]{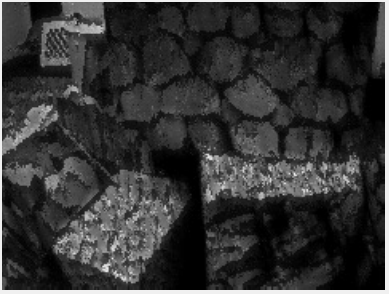}\\
\hspace{-0.25 cm}
(e) E2VID \cite{Rebecq_2019_CVPR}
&(f) mEDI \\
\end{tabular}
\end{center}
\vspace{-4mm}
\caption{\label{fig:manually} \em An example of our reconstruction result using different methods to estimate $c$, on a real sequence from the \emph{Event-Camera Dataset}~\cite{mueggler2017event}.
(a) The blurred image.
(b) Deblurring result of \cite{Tao_2018_CVPR}.
(c) Our result where $c$ is chosen by manual inspection.
(d) Our result where $c$ is computed automatically by our proposed energy minimization Eq.~\eqref{eq:solveK}.
\rc{(e) Reconstruction results of Rebecq \etal \cite{Rebecq_2019_CVPR} from only events. The temporal resolution of the reconstructed video is around $\times 8$ times higher than the original videos' based on their default settings.
(f) Our mEDI result where the temporal resolution is the same as (e).
}}
\end{figure}

\section{Optimization}
The unknown contrast threshold $c$ represents the minimum change in log intensity required to trigger an event.
With an appropriate $c$ in Eq.~\eqref{eq:mlogEDIM}, we can generate a sequence of sharper images.
Here, we propose two different methods to estimate the unknown variable $c$, which are manually chosen and automatically optimized by our approach.

\begin{table*}
\caption{\em Quantitative comparisons on the Synthetic dataset \cite{Nah_2017_CVPR}. The provided videos are able to generate not only blurred images but also event data. All methods are tested under the same blur condition, where methods \cite{Nah_2017_CVPR,Jin_2018_CVPR,Tao_2018_CVPR,Zhang_2018_CVPR} use GoPro dataset \cite{Nah_2017_CVPR} to train their models. Jin \cite{Jin_2018_CVPR} achieves their best performance when the image is down-sampled to 45\% mentioned in their paper. In this dataset, blurry images are generated by averaging every 11 frames, and treat the middle clean one (the 6$^{th}$ frame) as the ground truth. The top part in this figure aims to compare with deblurring methods and only the blurry image (the 6$^{th}$ frame) is evaluated. The bottom part shows the measures of whole reconstructed videos.}
\vspace{-5mm}
\begin{center}
\label{all_all}
\resizebox{\textwidth}{!}{
\begin{tabular}{c|c|c|c|c|c|c|c|c|c}
\hline
\multicolumn{10}{c}{Average result of the deblurred {\bf images} on dataset\cite{Nah_2017_CVPR}}  \\ \hline
            & Pan \cite{pan2017deblurring} & Sun \cite{sun2015learning}  & Gong \cite{gong2017motion}  & Jin \cite{Jin_2018_CVPR} & Tao \cite{Tao_2018_CVPR} & Zhang \cite{Zhang_2018_CVPR} & Nah \cite{Nah_2017_CVPR}  & EDI\cite{Pan_2019_CVPR} & mEDI  \\ \hline
PSNR(dB)  & 23.50  & 25.30  & 26.05   & 26.98    & 30.26     & 29.18   & 29.08    & 29.06  & {\bf 30.29} \\ \hline
SSIM      & 0.8336 & 0.8511 & 0.8632  & 0.8922   & 0.9342    & 0.9306  & 0.9135   & {\bf{0.9430}} & 0.9194  \\  \hline 
\multicolumn{10}{c}{Average result of the reconstructed {\bf videos} on dataset\cite{Nah_2017_CVPR}}  \\ \hline
\multicolumn{1}{c|}{}     & \multicolumn{2}{c|}{Baseline 1 \cite{Tao_2018_CVPR} + \cite{Scheerlinck18accv}} & \multicolumn{2}{c|}{Baseline 2 \cite{Scheerlinck18accv} + \cite{Tao_2018_CVPR}} & \multicolumn{2}{c|}{Scheerlinck \etal \cite{Scheerlinck18accv}} & \multicolumn{1}{c|}{Jin \etal \cite{Jin_2018_CVPR}}    & EDI\cite{Pan_2019_CVPR} & mEDI\\ \hline
\multicolumn{1}{c|}{PSNR(dB)} & \multicolumn{2}{c|}{25.52}          & \multicolumn{2}{c|}{26.34}          & \multicolumn{2}{c|}{25.84}           & \multicolumn{1}{c|}{25.62}  & 28.49   & {\bf{28.83 }}\\ \hline
\multicolumn{1}{c|}{SSIM} & \multicolumn{2}{c|}{0.7685}          & \multicolumn{2}{c|}{0.8090}          & \multicolumn{2}{c|}{0.7904}           & \multicolumn{1}{c|}{0.8556} & {\bf{0.9199 }} & 0.9098 \\ \hline
\end{tabular}
}
\end{center}
\end{table*}
\begin{figure*}[!ht]
\begin{center}
\begin{tabular}{cccccc}
\multicolumn{2}{c}{
\hspace{-0.3 cm}
\begin{tabular}{cc}
\includegraphics[height=0.135\textwidth]{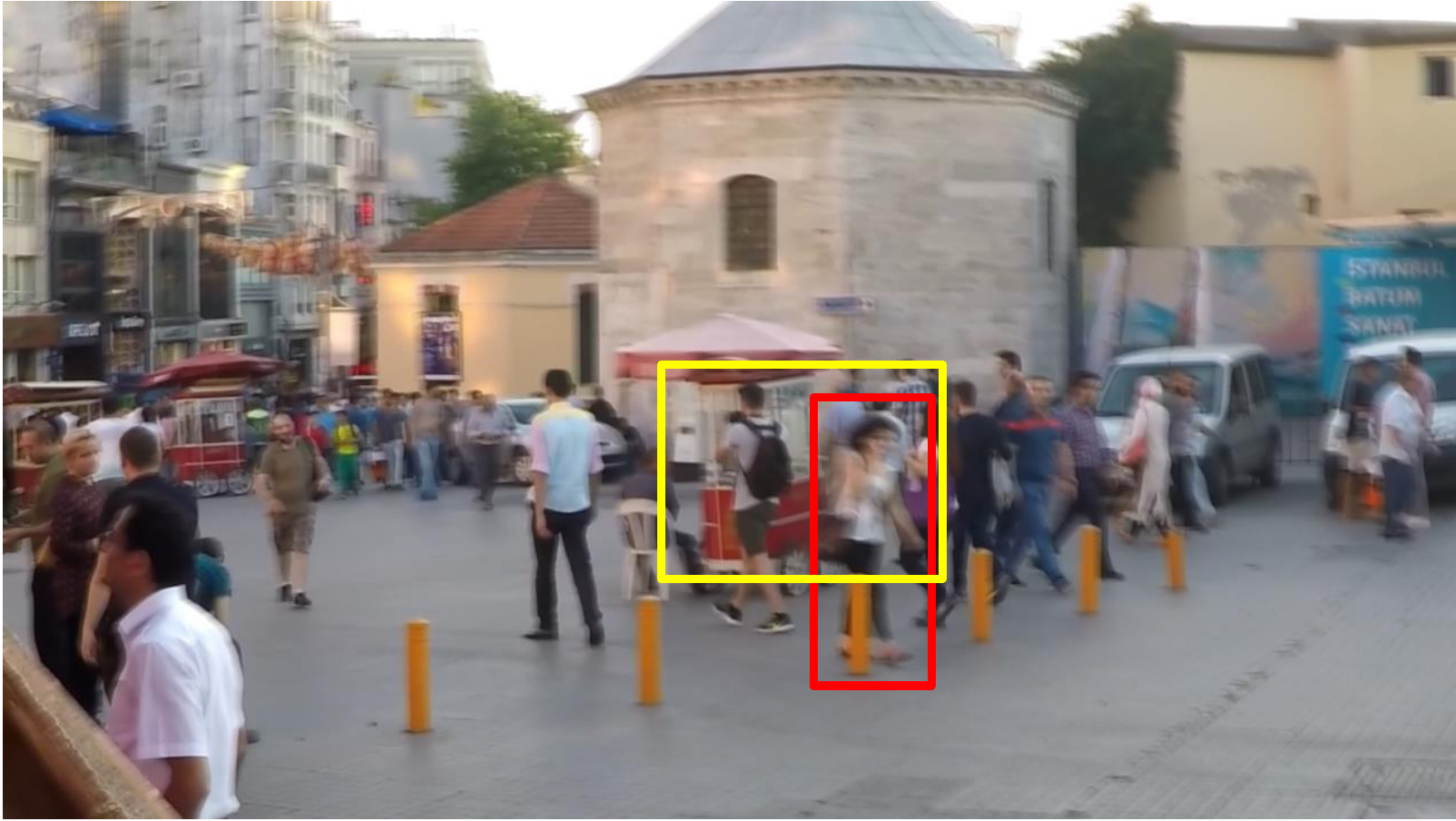}
\hspace{-0.4 cm}
&\includegraphics[height=0.135\textwidth]{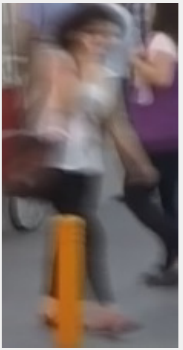}\\
\multicolumn{2}{c}{(a) The blurred Image}
\end{tabular}
}
&\multicolumn{2}{c}{
\hspace{-0.7 cm}
\begin{tabular}{cc}
\includegraphics[height=0.135\textwidth]{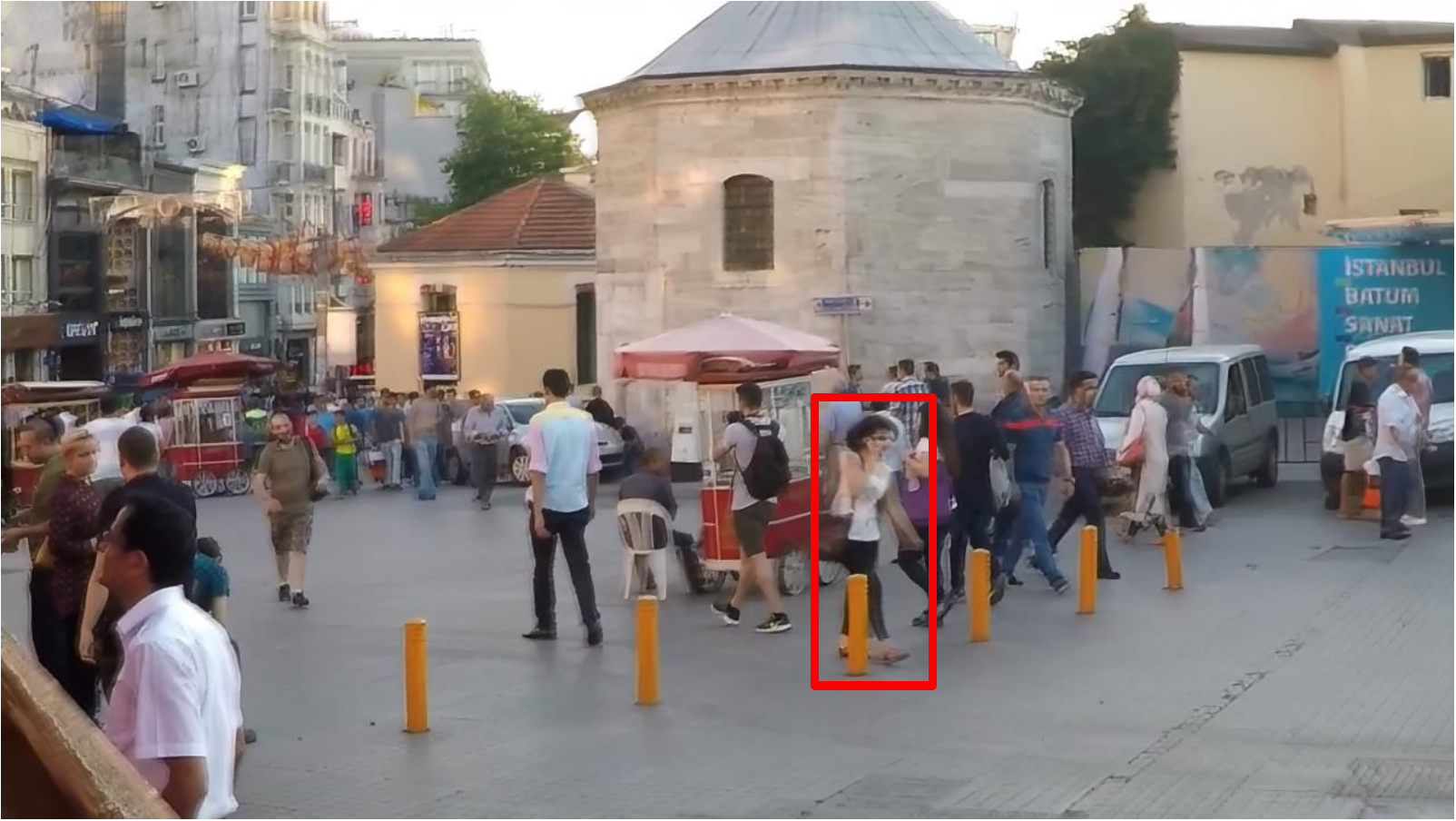}
\hspace{-0.4 cm}
&\includegraphics[height=0.135\textwidth]{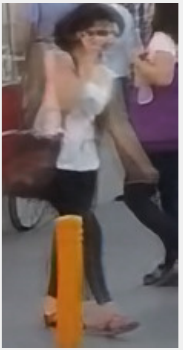}\\
\multicolumn{2}{c}{(b) Jin \etal \cite{Jin_2018_CVPR}}
\end{tabular}
}
&\multicolumn{2}{c}{
\hspace{-0.75 cm}
\begin{tabular}{cc}
\includegraphics[height=0.135\textwidth]{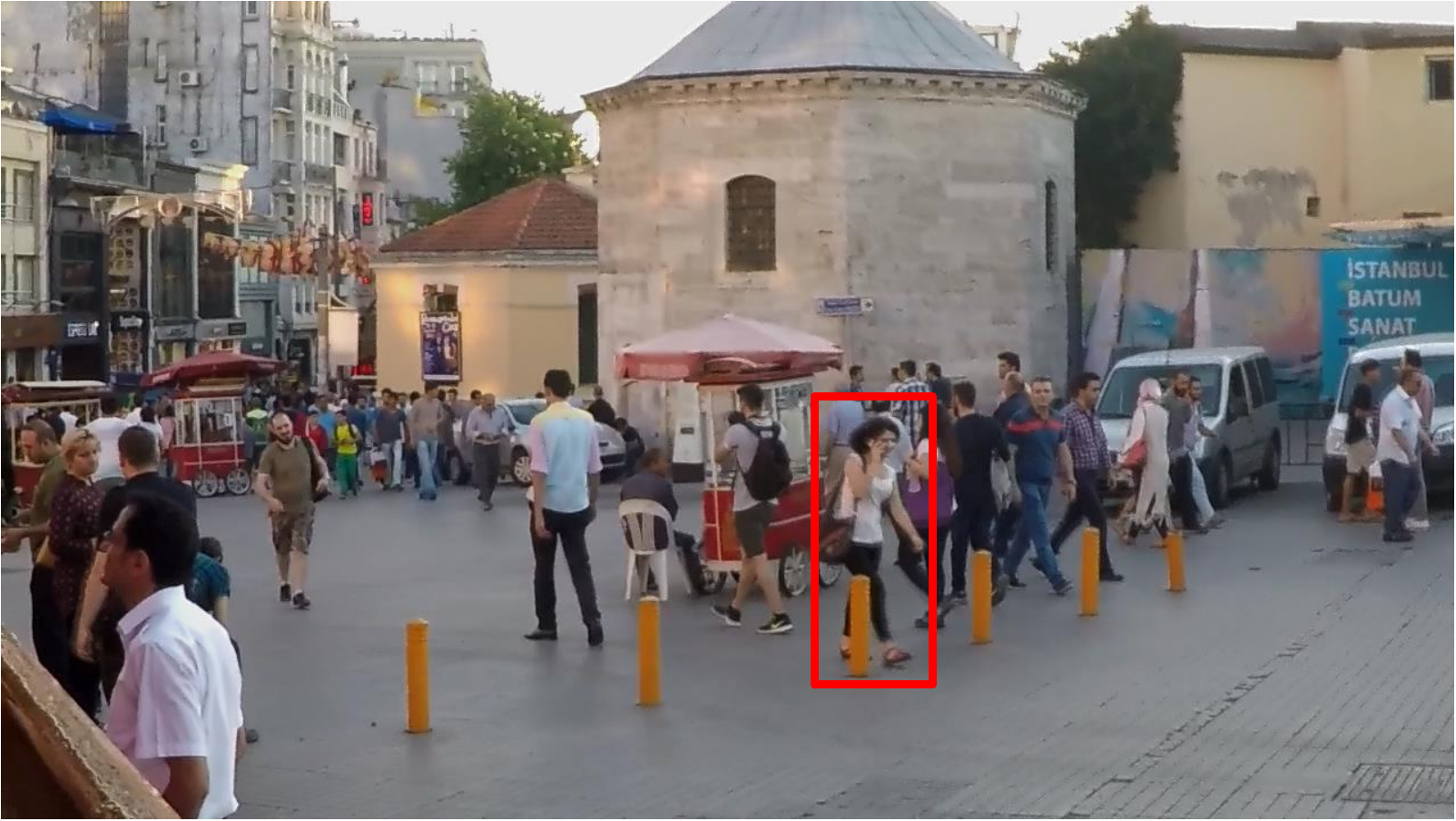}
\hspace{-0.4 cm}
&\includegraphics[height=0.135\textwidth]{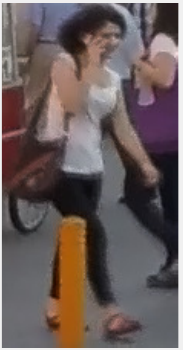}\\
\multicolumn{2}{c}{(c) Ours}
\end{tabular}
}\\
\multicolumn{6}{l}{
\hspace{-0.3 cm}
\begin{tabular}{ccccccc}
\includegraphics[height=0.136\textwidth]{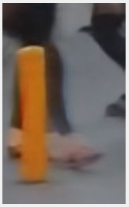}
\hspace{ 0.635 cm}
&\includegraphics[height=0.136\textwidth]{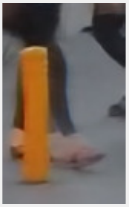}
\hspace{ 0.635 cm}
&\includegraphics[height=0.136\textwidth]{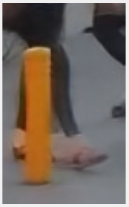}
\hspace{ 0.635 cm}
&\includegraphics[height=0.136\textwidth]{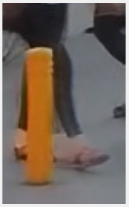}
\hspace{ 0.635 cm}
&\includegraphics[height=0.136\textwidth]{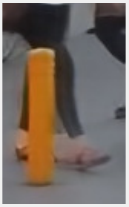}
\hspace{ 0.635 cm}
&\includegraphics[height=0.136\textwidth]{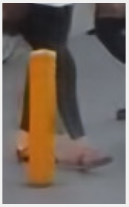}
\hspace{ 0.635 cm}
&\includegraphics[height=0.136\textwidth]{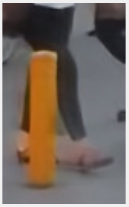}\\
\multicolumn{7}{c}{(d) The Reconstructed Video of \cite{Jin_2018_CVPR}}\\
\end{tabular}
}\\
\multicolumn{6}{l}{
\hspace{-0.3 cm}
\begin{tabular}{ccccccccccc}
\includegraphics[height=0.133\textwidth]{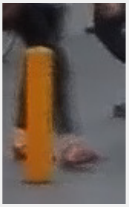}
\hspace{-0.400 cm}
&\includegraphics[height=0.133\textwidth]{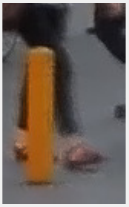}
\hspace{-0.400 cm}
&\includegraphics[height=0.133\textwidth]{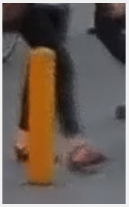}
\hspace{-0.400 cm}
&\includegraphics[height=0.133\textwidth]{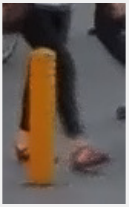}
\hspace{-0.400 cm}
&\includegraphics[height=0.133\textwidth]{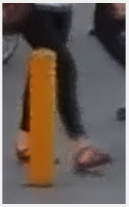}
\hspace{-0.400 cm}
&\includegraphics[height=0.133\textwidth]{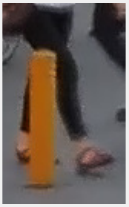}
\hspace{-0.400 cm}
&\includegraphics[height=0.133\textwidth]{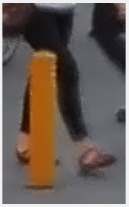}
\hspace{-0.400 cm}
&\includegraphics[height=0.133\textwidth]{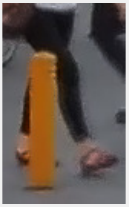}
\hspace{-0.400 cm}
&\includegraphics[height=0.133\textwidth]{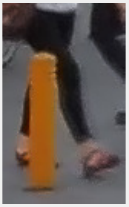}
\hspace{-0.400 cm}
&\includegraphics[height=0.133\textwidth]{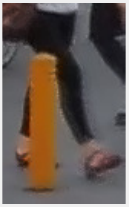}
\hspace{-0.400 cm}
&\includegraphics[height=0.133\textwidth]{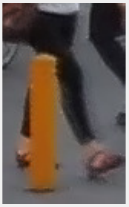}\\
\multicolumn{11}{c}{(e) The Reconstructed Video of mEDI}\\
\end{tabular}
}\\
\multicolumn{3}{c}{ 
\hspace{-0.3 cm}
\begin{tabular}{ccc}
\includegraphics[height=0.114\textwidth]{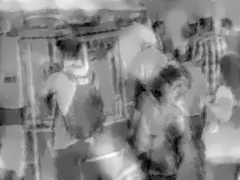}
\hspace{-0.35 cm}
&\includegraphics[height=0.114\textwidth]{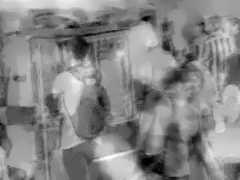}
\hspace{-0.35 cm}
&\includegraphics[height=0.114\textwidth]{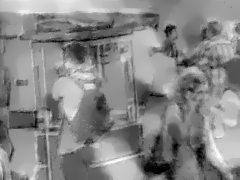}\\
\multicolumn{3}{c}{ (f) Reinbacher \etal \cite{Reinbacher16bmvc}}\\
\end{tabular}
}
&\multicolumn{3}{c}{ 
\hspace{-0.55 cm}
\begin{tabular}{ccc}
\includegraphics[height=0.114\textwidth]{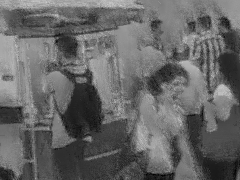}
\hspace{-0.35 cm}
&\includegraphics[height=0.114\textwidth]{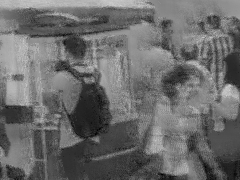}
\hspace{-0.35 cm}
&\includegraphics[height=0.114\textwidth]{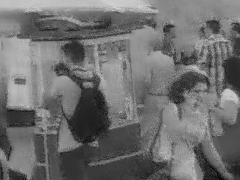}\\
\multicolumn{3}{c}{ (g) Scheerlinck \etal \cite{Scheerlinck18accv}}\\
\end{tabular}
}\\
\end{tabular}
\end{center}
\vspace{-4mm}
\caption{\em \label{fig:VideoGoPro} An example of the reconstructed result on our synthetic event dataset based on the GoPro dataset~\cite{Nah_2017_CVPR}. \cite{Nah_2017_CVPR} provides videos to generate blurred images and event data. 
(a) The blurred image. The red close-up frame is for (b)-(e), the yellow close-up frame is for (f)-(g). 
(b) The deblurring result of Jin \etal \cite{Jin_2018_CVPR}. 
(c) Our deblurring result. 
(d) The crop of their reconstructed images and the frame number is fixed at 7. Jin \etal \cite{Jin_2018_CVPR} uses the GoPro dataset added with 20 scenes as training data and  their model is supervised by 7 consecutive sharp frames.
(e) The crop of our reconstructed images. 
(f) The crop of Reinbacher \cite{Reinbacher16bmvc} reconstructed images from only events.
(g) The crop of Scheerlinck \cite{Scheerlinck18accv} reconstructed image, they use both events and the intensity image.
For (e)-(g), the shown frames are the chosen examples, where the length of the reconstructed video is based on the number of events. (Best viewed on screen).
}
\end{figure*}


\subsection{Manually Chosen $c$}
According to our {\bf mEDI} model in Eq.~\eqref{eq:mlogEDIM}, given a value for $c$, we obtain sharp images.
Therefore, we develop a method for deblurring by manually inspecting the visual effect of the deblurred image. 
In this way, we incorporate human perception into the reconstruction loop and the deblurred images should satisfy human observation.
In Fig.~\ref{fig:eventsample} and \ref{fig:manually}, we give examples for manually chosen results on our dataset, and the \emph{Event-Camera Dataset}~\cite{mueggler2017event}. 

\vspace{-1mm}
\subsection{Automatically Chosen $c$}

To automatically find the best $c$, we need to build an evaluation metric (energy function) that can evaluate the quality of the deblurred image $\vL_i(t)$. Different from our EDI that including regularization terms (with extra weight parameters) in the energy function, we develop a simple yet effective optimization solution. More specifically, we adopt the Fibonacci sequence search to solve the optimization which significantly reduces the computational complexity.

\subsubsection{Energy function}


The values on the right-hand side of Eq.~\eqref{eq:mlogEDIM} depend on an unknown parameter $c$. In particular, we write 
\begin{equation}
\begin{split}
b_i& =  c \, \int_{f_i}^t e(s) ds\\
a_i& = \log \left(\frac{1}{T}\int_{f_i-T/2}^{f_i+T/2} \exp(c \,\vE(t)) dt\right).
\end{split}
\end{equation}

\begin{figure}[t]
 \begin{center}
 \includegraphics[width=0.475\textwidth]{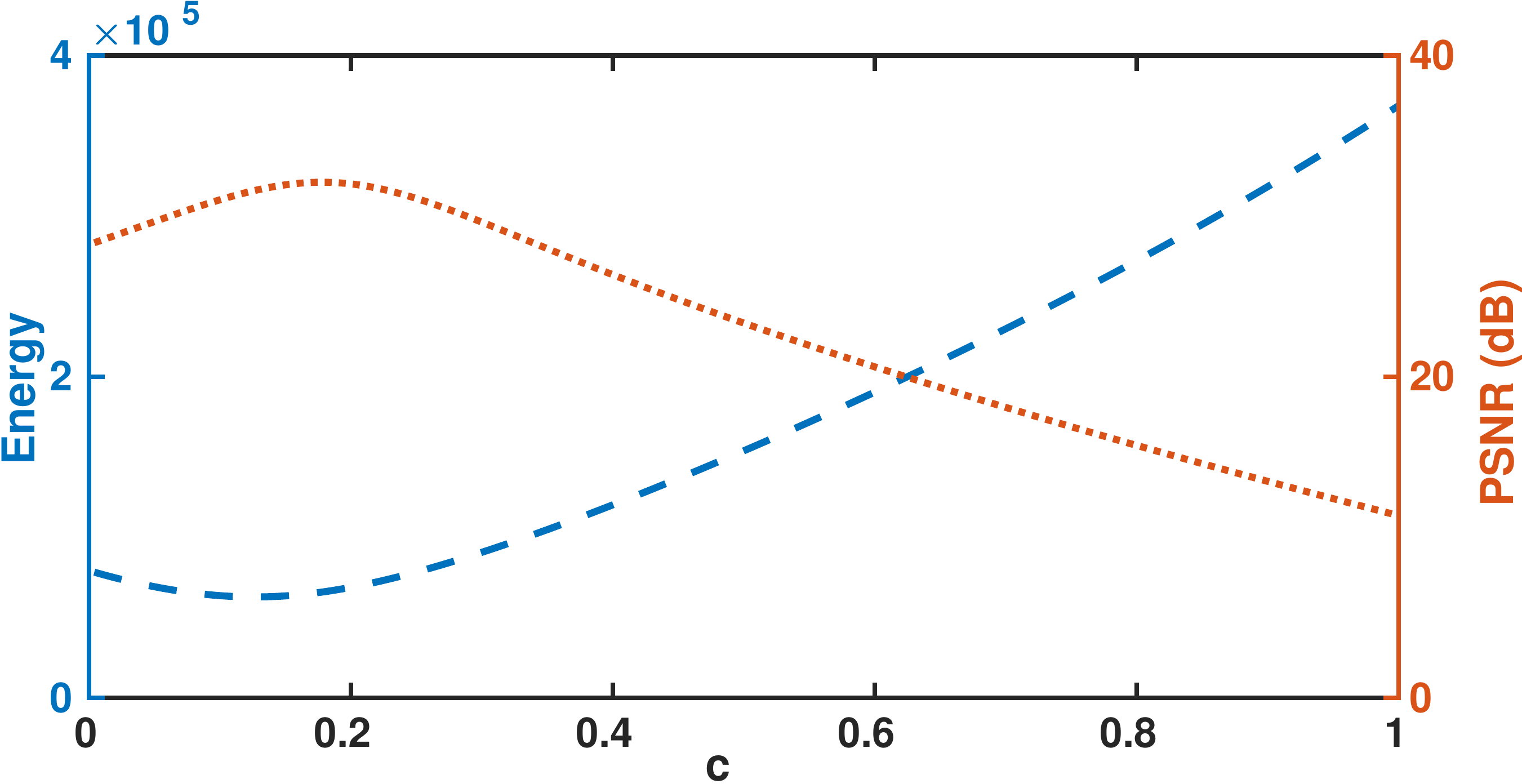}
 \end{center}
 \vspace{-4mm}
 \caption{\em \label{fig:deltaVsblur} Deblurring performance plotted against the value of $c$.
 The image is clearer with higher PSNR value.}
 \end{figure}

Given $c$, ${{x}_i}$ can be solved by $\tt{LU}$ decomposition in Sec. \ref{sec:LU}. Subsequently, from Eq. \eqref{eq:mlogEDIM} the blur image ${\vB}_i$ can be computed. 
\begin{equation}
\widetilde{\vB}_i(c) = {x}_i+a_i
\end{equation}
 Here, we use $\vB_i(c)$ to present the blurred image $\vB_i$ with different $c$. In this case, the optimal $c$ can be estimated by solving Eq.~\eqref{eq:solveK},
%
\begin{equation}\label{eq:solveK}
\min_{c} ||\vB_i(c)-\vB||_2^2.
\end{equation}
Examples show that as a function of $c$, the residual error in solving the equations is not convex.
However, in most cases (empirically) it seems to be convex, or at least it has a single minimum (See Fig.~\ref{fig:deltaVsblur} for an example).

\subsubsection{Fibonacci search}

Finding the minimum of a function along a single line is easy if that function has a single minimum. 
In the case of least-squares minimization problems, various strategies for determining the line-search direction are currently used, such as conjugate gradient methods, gradient descent, and the  Levenberg-Marquardt method.
%
When the function has only one stationary point, the maximum/minimum, and when it depends on a single variable in a finite interval, the most efficient way to find the maximum is based on the Fibonacci numbers.
The procedure, now known widely as `Fibonacci search', was discovered and shown optimal in a minimax sense by Kiefer \cite{kiefer1953sequential, press1988numerical}. 

In this work, we use Fibonacci search for the value of $c$ that gives the least error.
~In Fig.~\ref{fig:deltaVsblur}, we illustrate the clearness of the reconstructed image (in PSNR value) as a function of the value of $c$. 
As demonstrated in the figure, our proposed reconstruction metric could locate/identify the best deblurred image with peak PSNR properly. 

\rcf{
In our proposed method, we assume that $c_+ = c_-$ and use a global $c$ based on the following reasons:
\begin{enumerate}[1)]
    \item As illustrated in Fig.~\ref{fig:deltaVsblur} our deblurring performance has a relatively flat crest against different values of $c$. Experimental results demonstrated that the quality of our reconstructed videos is robust to the estimation of $c$ within a certain range. 
    \item 
    We have conducted the experiments with $c_+ \neq c_-$, namely, optimising two variables in our formulation. 
    We observed that the improvement on PSNR is less than $0.1$dB in comparison to the results of optimising a global $c$. However, the computational complexity increases from $\mathcal{O}(n)$ to $\mathcal{O}(n^2)$.
\end{enumerate}
    Therefore, we believe it is worthy of trading off between computational simplicity and performance accuracy. }
\begin{figure*}
\begin{center}
\begin{tabular}{cccc}
\includegraphics[width=0.225\textwidth]{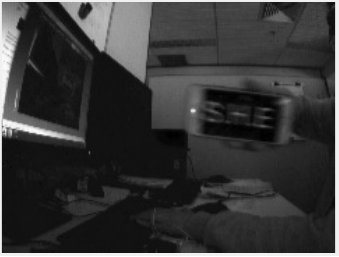}
&\includegraphics[width=0.225\textwidth]{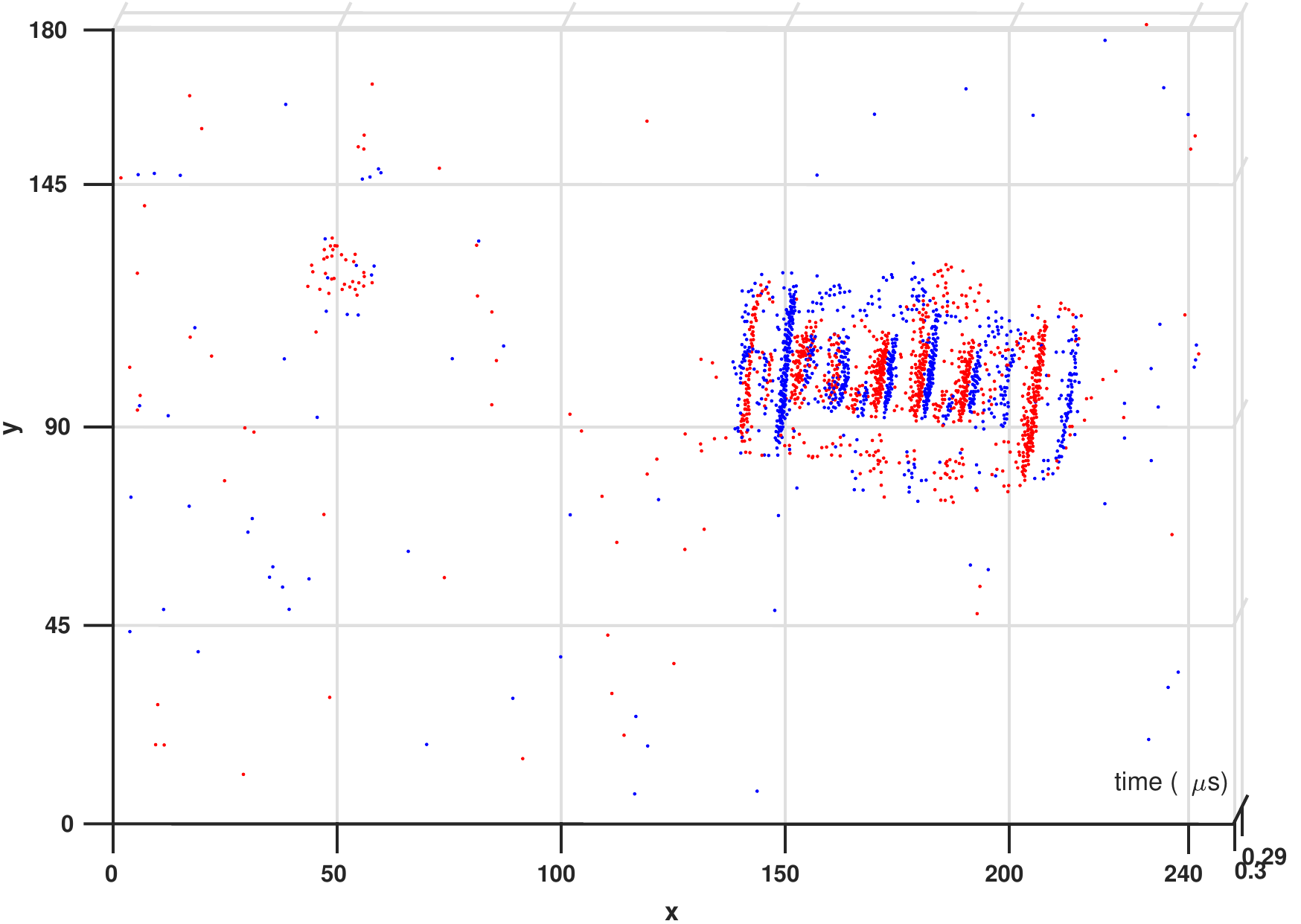}
&\includegraphics[width=0.225\textwidth]{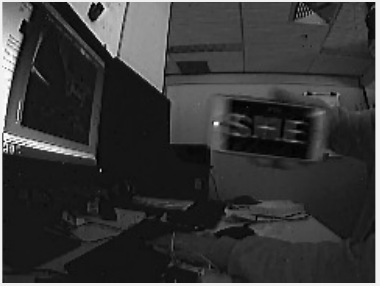}
&\includegraphics[width=0.225\textwidth]{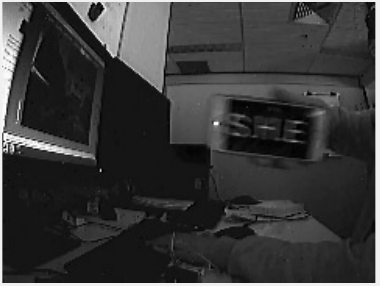}\\
\includegraphics[width=0.225\textwidth]{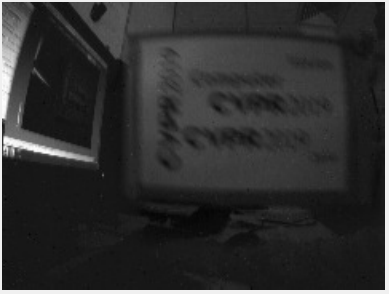}
&\includegraphics[width=0.225\textwidth]{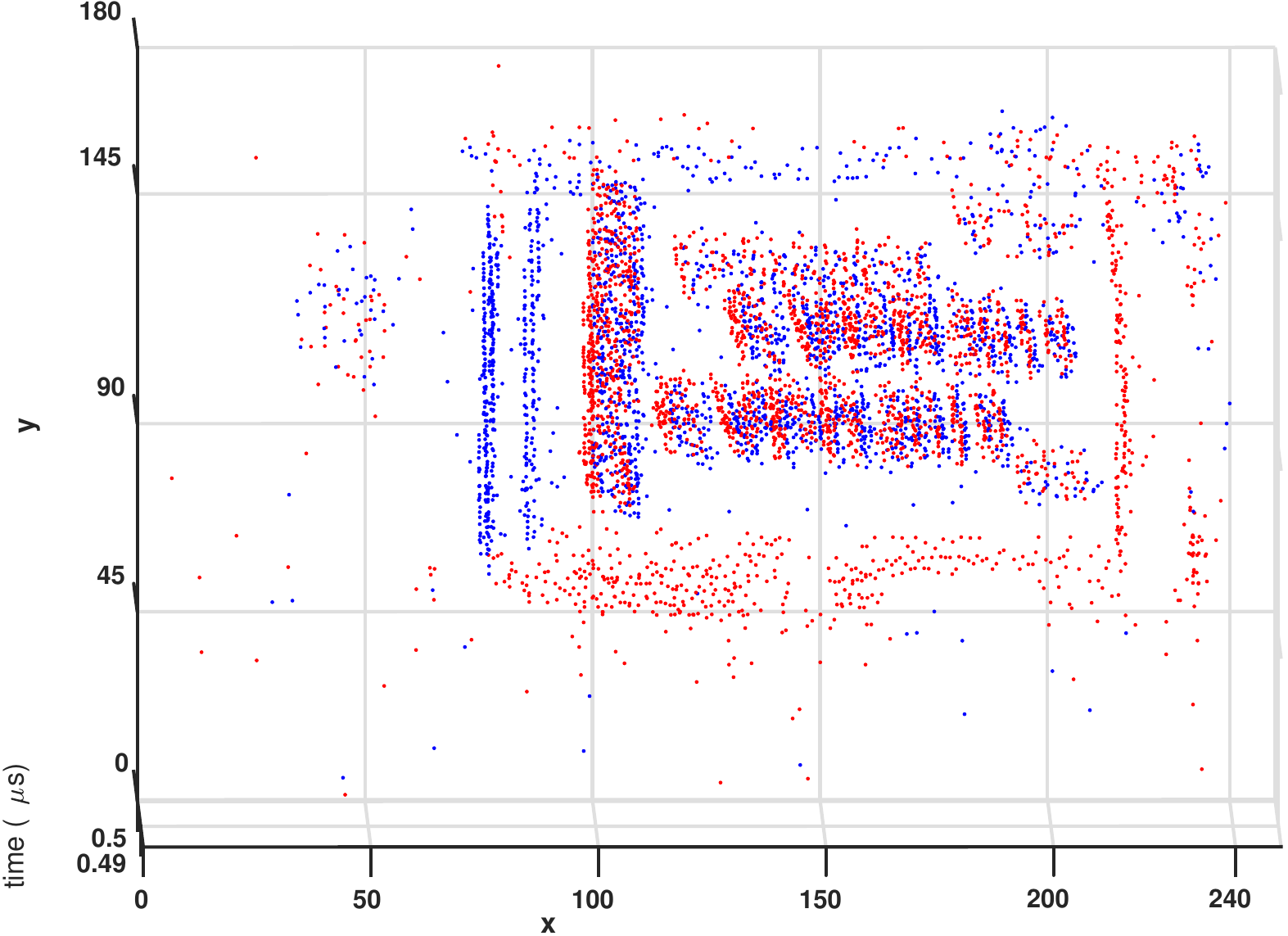}
&\includegraphics[width=0.225\textwidth]{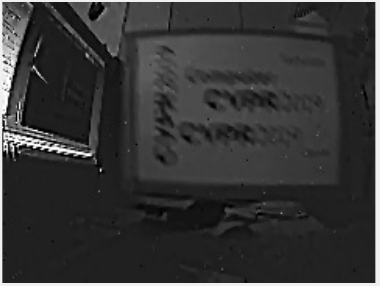}
&\includegraphics[width=0.225\textwidth]{./aReal/0013_pan.pdf}\\
(a) The Blurred Image  
&(b) The Event
&(c) Pan \etal \cite{pan2017deblurring}
&(d) yan \etal \cite{yan2017image}\\
\includegraphics[width=0.225\textwidth]{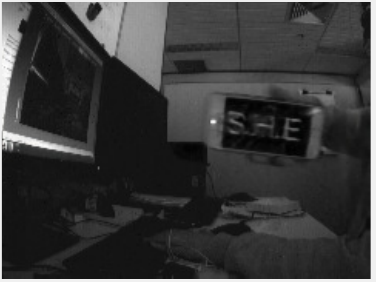}
&\includegraphics[width=0.225\textwidth]{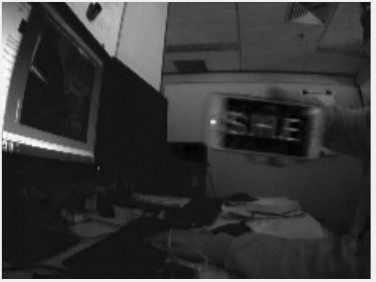}
&\includegraphics[width=0.225\textwidth]{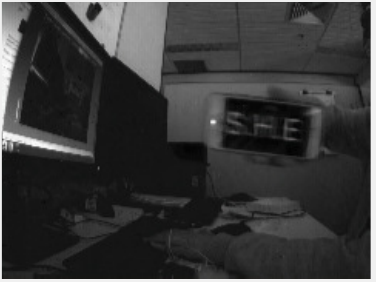}
&\includegraphics[width=0.225\textwidth]{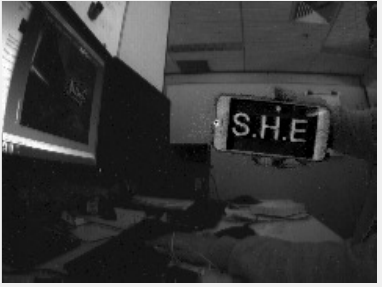}\\
\includegraphics[width=0.225\textwidth]{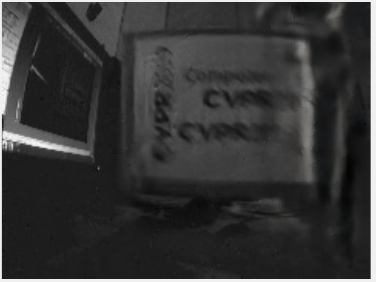}
&\includegraphics[width=0.225\textwidth]{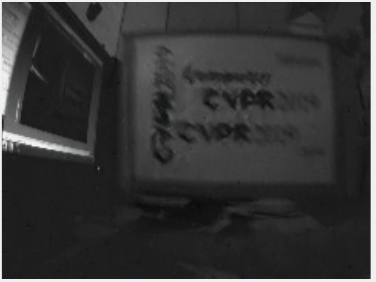}
&\includegraphics[width=0.225\textwidth]{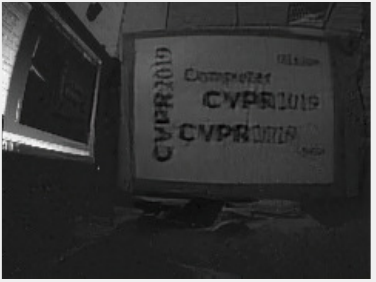}
&\includegraphics[width=0.225\textwidth]{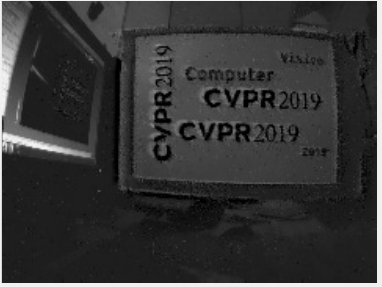}\\
(e) Tao \etal \cite{Tao_2018_CVPR}
&(f) Nah \etal \cite{Nah_2017_CVPR}
&(g) Jin \etal \cite{Jin_2018_CVPR} 
&(h) Our EDI \cite{Pan_2019_CVPR}\\
\includegraphics[width=0.225\textwidth]{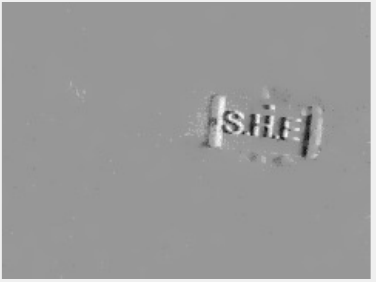}
&\includegraphics[width=0.225\textwidth]{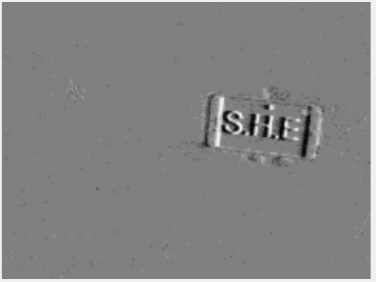}
&\includegraphics[width=0.225\textwidth]{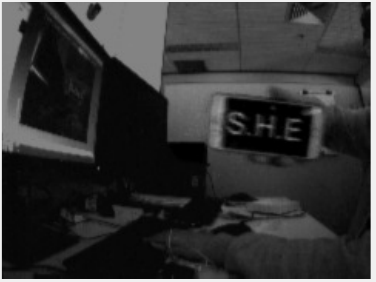}
&\includegraphics[width=0.225\textwidth]{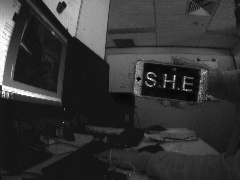}\\
\includegraphics[width=0.225\textwidth]{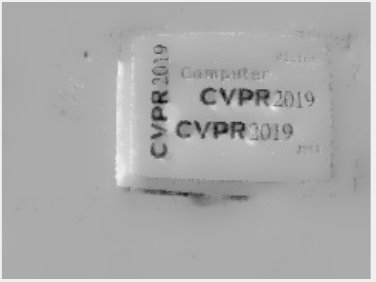}
&\includegraphics[width=0.225\textwidth]{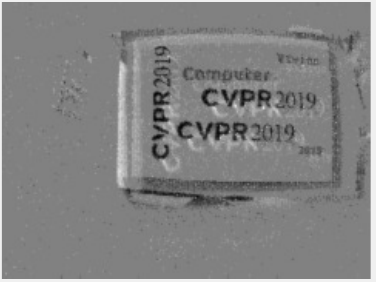}
&\includegraphics[width=0.225\textwidth]{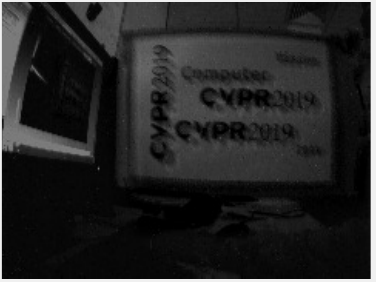}
&\includegraphics[width=0.225\textwidth]{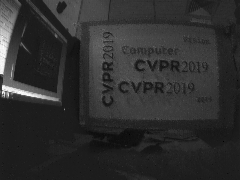}\\
(i) Reinbacher \etal \cite{Reinbacher16bmvc}  
&(j)  {\small\begin{tabular}[c]{@{}c@{}}Scheerlinck \etal \cite{Scheerlinck18accv}\\ (events only)\end{tabular}}
&(k) Scheerlinck \etal \cite{Scheerlinck18accv}
&(l) Our mEDI\\
\end{tabular}
\end{center}
 \caption{\em \label{fig:Real} Examples of reconstruction result on our real {\it blur event dataset} in low lighting and complex dynamic conditions
(a) Input blurred images. 
(b) The event information. 
(c) Deblurring results of \cite{pan2017deblurring}. 
(d) Deblurring results of \cite{yan2017image}. 
(e) Deblurring results of \cite{Tao_2018_CVPR}. 
(f) Deblurring results of \cite{Nah_2017_CVPR}.  
(g) Deblurring results of \cite{Jin_2018_CVPR} and they use video as training data. 
(h) Reconstruction result of \cite{Pan_2019_CVPR} from combining events and frames.
(i) Reconstruction result of \cite{Reinbacher16bmvc} from only events. 
(j)-(k) Reconstruction results of \cite{Scheerlinck18accv}, (j) from only events, (k) from combining events and frames. 
(l) Our reconstruction result. 
Results in (c)-(g) show that real high dynamic settings and low light conditions are still challenging in the deblurring area. Results in (i)-(j) show that while intensity information of a scene is still retained with an event camera recording, color, and delicate texture information cannot be recovered. (Best viewed on screen).
}
\end{figure*}

\section{Experiment}
In all of our experiments, unless otherwise specified, the parameter $c$ for reconstructing images is chosen automatically by our optimization process.

\vspace{-1mm}
\subsection{Experimental Setup}

\noindent{\bf{Synthetic dataset.}} 
In order to provide a quantitative comparison, we build a synthetic dataset based on the GoPro blur dataset \cite{Nah_2017_CVPR}.
It supplies ground truth videos which are used to generate the blurred images. Similarly, we employ the ground-truth images to generate event data based on the methodology of \emph{event camera model}. 
\rcf{In this GoPro dataset, we did not notice obvious rolling shutter artifacts because images in this dataset were requested to be captured with low speed of camera motions for providing ground-truth latent sharp images.}

\vspace{0.5mm}
\noindent{\bf{Real dataset.}} 
We evaluate our method on a public Event-Camera dataset \cite{mueggler2017event}, which provides a collection of sequences captured by the event camera for high-speed robotics.
Furthermore, we present our real {\it blur event dataset}, 
where each real sequence is captured with the DAVIS240~\cite{brandli2014240} under different conditions, such as indoor, outdoor scenery, low lighting conditions, and different motion patterns (\eg, camera shake, objects motion) that naturally introduce motion blur into the APS intensity images. We also evaluate our method on a newly published Color Event Camera Dataset (CED)~\cite{scheerlinck2019ced} built with DAVIS346 Red Color sensor. \rc{They present an extension of the event  simulator ESIM~\cite{rebecq2018esim} that enables simulation of color events. }
\rcf{In contrast to GoPro cameras, event cameras, such as DAVIS, employ global shutters, where an entire scene is captured at the same instant. Therefore, global shutter cameras, \eg, our event camera, do not have rolling shutter effects.}

\vspace{0.5mm}
\noindent{\bf{Implementation details.}} 
For all our real experiments, we use the DAVIS~\cite{brandli2014240} that shares photosensor array to simultaneously output events (DVS) and intensity images (APS).
The framework is implemented using MATLAB.
It takes around 1.5 seconds to process one image on a single i7 core running at 3.6 GHz.

\begin{figure}[t]
\begin{center}
\begin{tabular}{cccccc}
\hspace{-0.35 cm}
\includegraphics[width=0.152\textwidth]{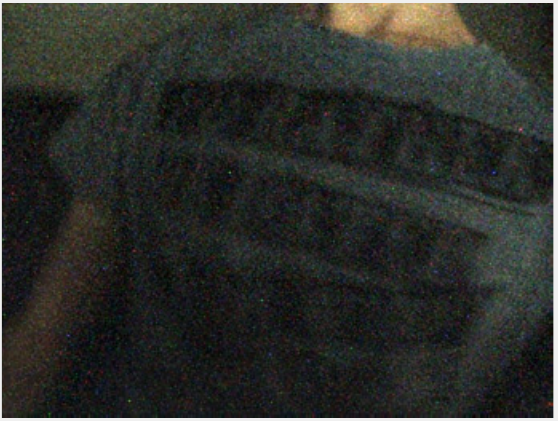}  
\hspace{-0.45 cm}
&\includegraphics[width=0.152\textwidth]{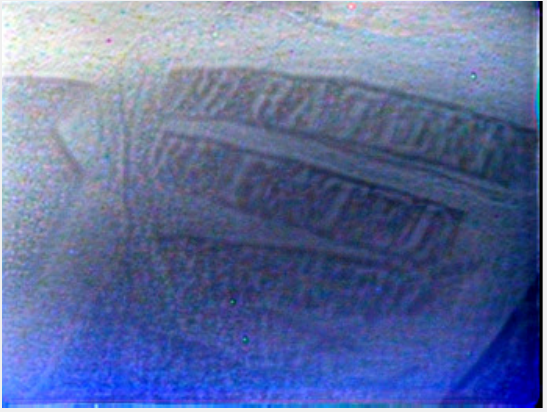} 
\hspace{-0.45 cm}
&\includegraphics[width=0.152\textwidth]{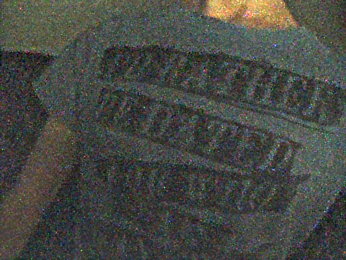}\\
\hspace{-0.35 cm}
\includegraphics[width=0.152\textwidth]{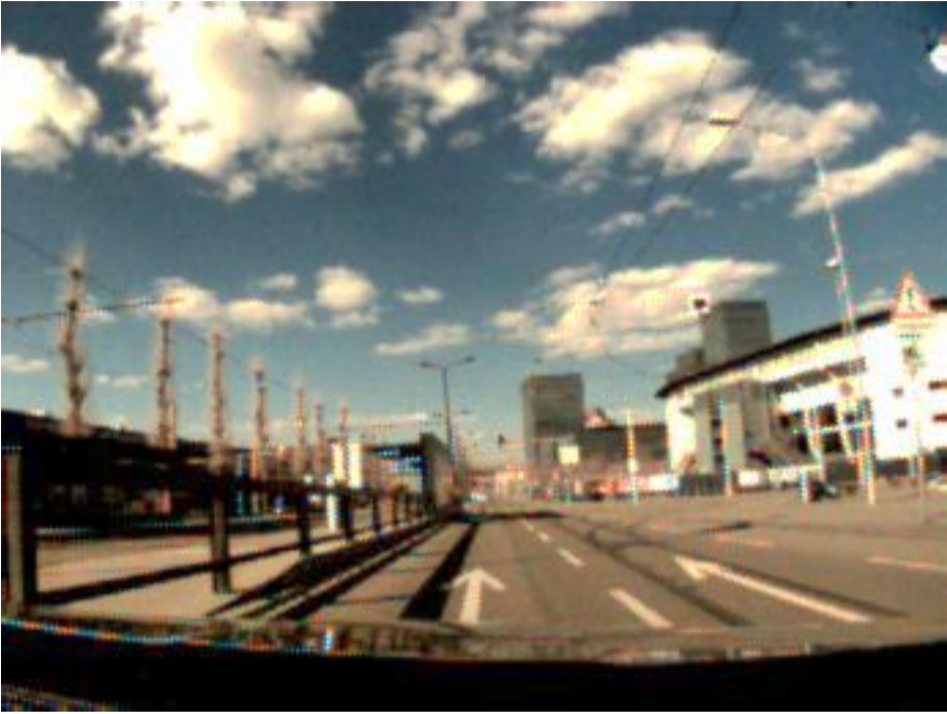}  
\hspace{-0.45 cm}
&\includegraphics[width=0.152\textwidth]{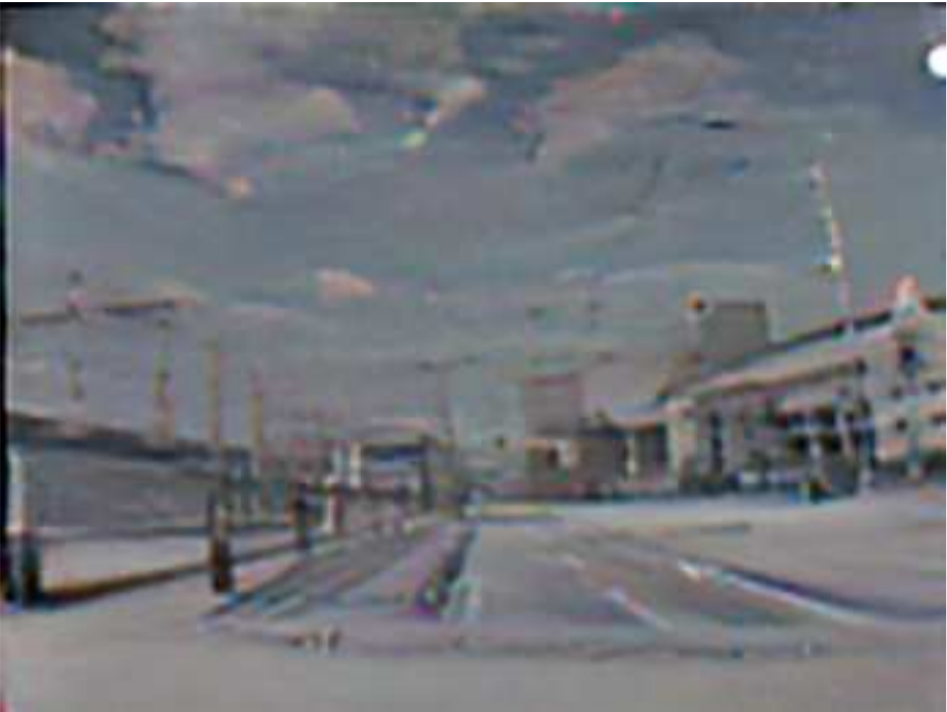} 
\hspace{-0.45 cm}
&\includegraphics[width=0.152\textwidth]{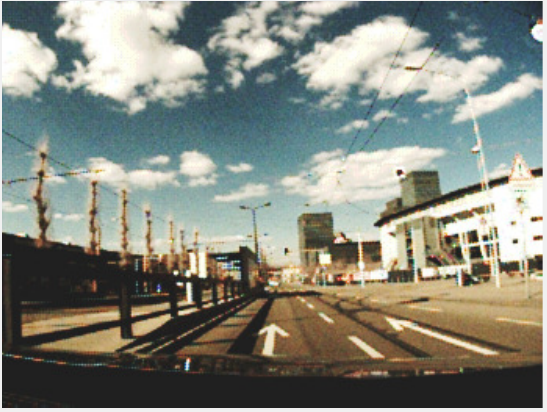}    \\
\hspace{-0.35 cm}
\includegraphics[width=0.152\textwidth]{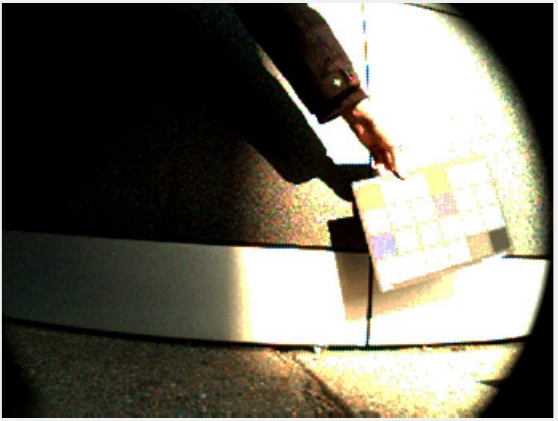}  
\hspace{-0.45 cm}
&\includegraphics[width=0.152\textwidth]{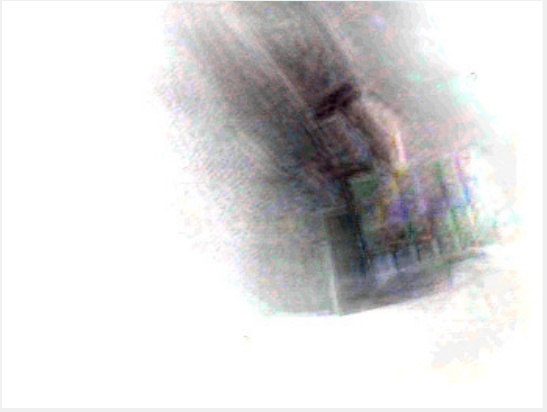} 
\hspace{-0.45 cm}
&\includegraphics[width=0.152\textwidth]{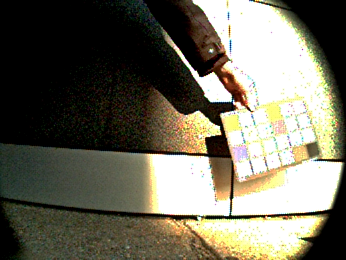}\\
\hspace{-0.35 cm}
\includegraphics[width=0.152\textwidth]{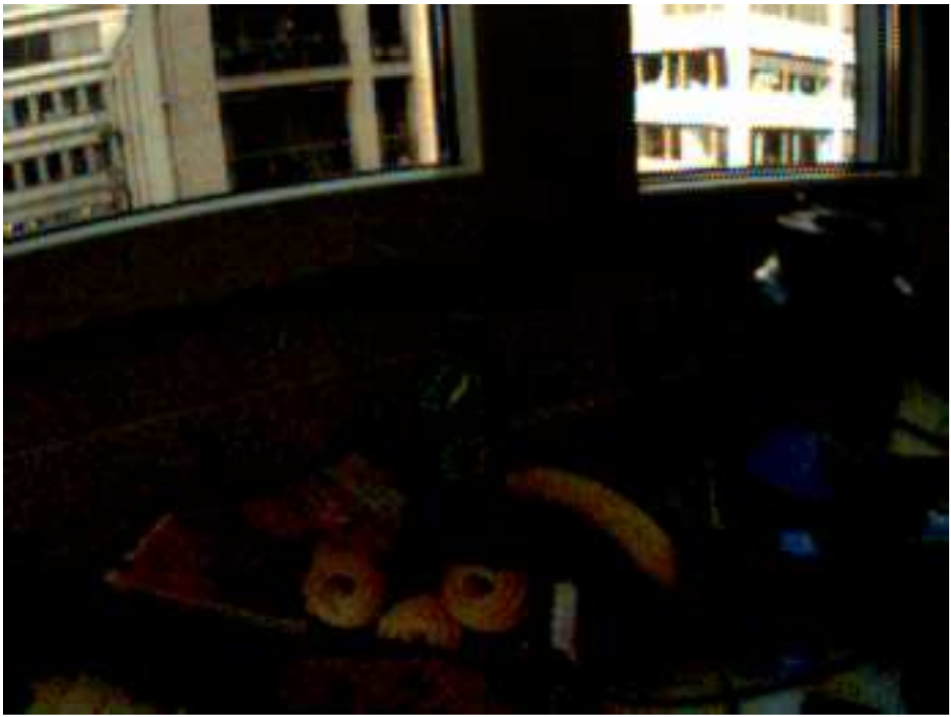}  
\hspace{-0.45 cm}
&\includegraphics[width=0.152\textwidth]{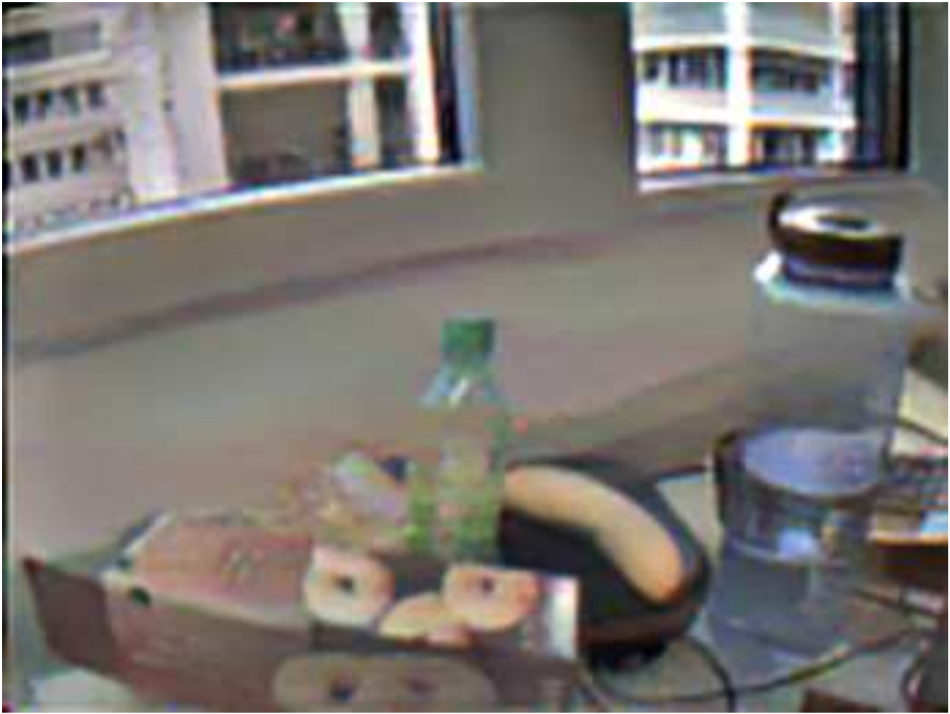} 
\hspace{-0.45 cm}
&\includegraphics[width=0.152\textwidth,height=0.1155\textwidth]{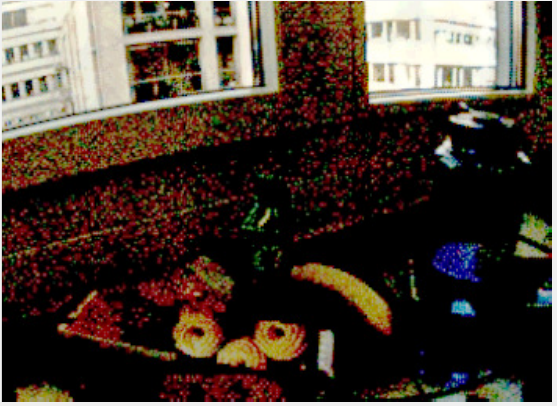}\\
\hspace{-0.35 cm}
(a) DAVIS frame
\hspace{-0.45 cm}
&(b) E2VID \cite{Rebecq_2019_CVPR}
\hspace{-0.45 cm}
&(c) Ours\\
\end{tabular}
\vspace{-4mm}
\end{center}
\caption{\label{fig:ced} \em 
\rc{An example of our reconstruction result on the color event camera dataset CED~\cite{scheerlinck2019ced}.
(a) The input image.
(b) Reconstruction results of Rebecq \etal \cite{Rebecq_2019_CVPR} from only events. The temporal resolution of the reconstructed video is around $\times 12$ times higher than the original videos' based on their default settings.
(c) Our mEDI result where the temporal resolution is the same as (b). 
\rcf{
From top to bottom, a scene with a low lighting condition, an outdoor scene, a scene with slow-moving objects (static background), and an HDR scene. Our mEDI model performs well in the top two rows, while E2VID is able to provide vivid color textures in the HDR scene. 
Note that our method focuses on reconstructing high-frame rate videos rather than changing the dynamic range of input videos.
In order to illustrate our detailed textures in the HDR scene, we employ an HDR enhancement method \cite{eilertsen2017hdr}. 
} 
}}
\end{figure}
\vspace{-1mm}
\subsection{Experimental Results}

We compare our proposed approach with state-of-the-art blind deblurring methods, including conventional deblurring methods \cite{pan2017deblurring,yan2017image}, deep based dynamic scene deblurring methods \cite{Nah_2017_CVPR,Jin_2018_CVPR,Tao_2018_CVPR,Zhang_2018_CVPR,sun2015learning}, and event-based image reconstruction methods \cite{Rebecq_2019_CVPR,Reinbacher16bmvc,Scheerlinck18accv}. Moreover, Jin \etal \cite{Jin_2018_CVPR} can restore a video from a single blurred image based on a deep network, where the middle frame in the restored odd-numbered sequence is the best.

To prove the effectiveness of our model, we show some baseline comparisons in Fig. \ref{fig:baseline} and Table \ref{all_all}. 
For baseline 1, we first apply a state-of-the-art deblurring method~\cite{Tao_2018_CVPR} to recover a sharp image, and then feed the recovered image as input to a reconstruction method~\cite{Scheerlinck18accv}.
For baseline 2, we first use the video reconstruction method~\cite{Scheerlinck18accv} to reconstruct a sequence of intensity images, then apply the deblurring method~\cite{Tao_2018_CVPR} to each frame.
As seen in Table~\ref{all_all}, our approach obtains higher PSNR and SSIM in comparison to both baseline 1 and baseline 2. This also implies that our approach better exploits the event data to not only recover sharp images but also reconstruct high frame rate videos.

In Table~\ref{all_all} and Fig.~\ref{fig:VideoGoPro}, we show quantitative and qualitative comparison 
on our synthetic dataset, respectively.
As indicated in Table~\ref{all_all}, our approach achieves the best performance on PSNR and competitive results on SSIM compared to state-of-the-art methods, and attains significant performance improvements on high-frame video reconstruction. 

\rcf{In Fig.~\ref{fig:multiimage}, Fig.~\ref{fig:comparecvprandnow} and Fig.~\ref{fig:ced}, we qualitatively compare our generated videos with state-of-the-art event-based image reconstruction methods \cite{Rebecq_2019_CVPR,Scheerlinck18accv,Pan_2019_CVPR}. Experimental results indicate that event-only methods work well on scenes of fast camera motions since the distribution of events has a wide coverage of scene content. Also, E2VID  \cite{Rebecq_2019_CVPR} is enabled to provide more vivid color textures in the HDR scene. However, for scenes with a static background or a slowly moving background/foreground, the reconstructed images by event-only methods will lose texture details in the areas without events. On the contrary, our `image and event' combined method achieves superior performance on scenes with high dynamic motions and works robustly even with static backgrounds and sparse events. 
}

We also report our reconstruction (and deblurring) results on real dataset, including text images and low-lighting images, in 
Fig.~\ref{fig:eventdeblur}, Fig.~\ref{fig:manually}, and Fig.~\ref{fig:Real}. 

Compared with state-of-the-art deblurring methods, our method achieves superior results. 
In comparison to existing event-based image reconstruction methods \cite{Reinbacher16bmvc, Scheerlinck18accv, Pan_2019_CVPR,Rebecq_2019_CVPR}, our reconstructed images are not only more realistic but also contain richer details. 
For more deblurring results and {\bf high-temporal resolution videos}, please visit our \href{https://github.com/panpanfei/Bringing-a-Blurry-Frame-Alive-at-High-Frame-Rate-with-an-Event-Camera}{home page}.

\section{Limitation}

Though event cameras record continuous, asynchronous streams of events that encode non-redundant information for our {\bf mEDI} model, there are still some limitations when doing reconstruction.

\begin{enumerate}[1)]
    \item Extreme lighting changes, such as suddenly turning on/off the light, moving from dark indoor scenes to outdoor scenes. \rc{The relatively low dynamic range of the intensity image might degrade the performance of our method in high dynamic scenes}; 
    \item Event error accumulation, such as noisy event data, small object motions with fewer events.
    Though we integrate over small time intervals from the centre of the exposure time to mitigate this error, accumulated noise can reduce the quality of reconstructed images. 
\end{enumerate}

\section{Conclusion}

In this paper, we have proposed a {\textbf{multiple Event-based Double Integral (mEDI)}} model to naturally connect intensity images and events recorded by an event camera (\rc{DAVIS}), which also takes the blur generation process into account.
In this way, our model can be used to not only recover the latent sharp images but also reconstruct intermediate frames at a high frame rate. We also propose a simple yet effective method to solve our {\bf mEDI} model.
Due to the simplicity of our optimization process, our method is efficient as well. Extensive experiments show that our method can generate high-quality, high frame-rate videos efficiently under different conditions, such as low lighting and complex dynamic scenes.


%



\ifCLASSOPTIONcompsoc
  \section*{Acknowledgments}
\else
  \section*{Acknowledgment}
\fi
This research was supported in part by the Australian Research Council through the ``Australian Centre of Excellence for Robotic Vision'' CE140100016, the Natural Science Foundation of China grants (61871325, 61420106007, 61671387, 61603303), National Key Research and Development Program of China under Grant 2018AAA0102803 and the Australian Research Council (ARC) grants (DE140100180, DE180100628, DP200102274).%

\ifCLASSOPTIONcaptionsoff
  \newpage
\fi

\bibliographystyle{IEEEtran}
\bibliography{pamibib}
\vspace{-5 mm}
\begin{IEEEbiography}[{\includegraphics[width=1in,height=1.25in,clip,keepaspectratio]{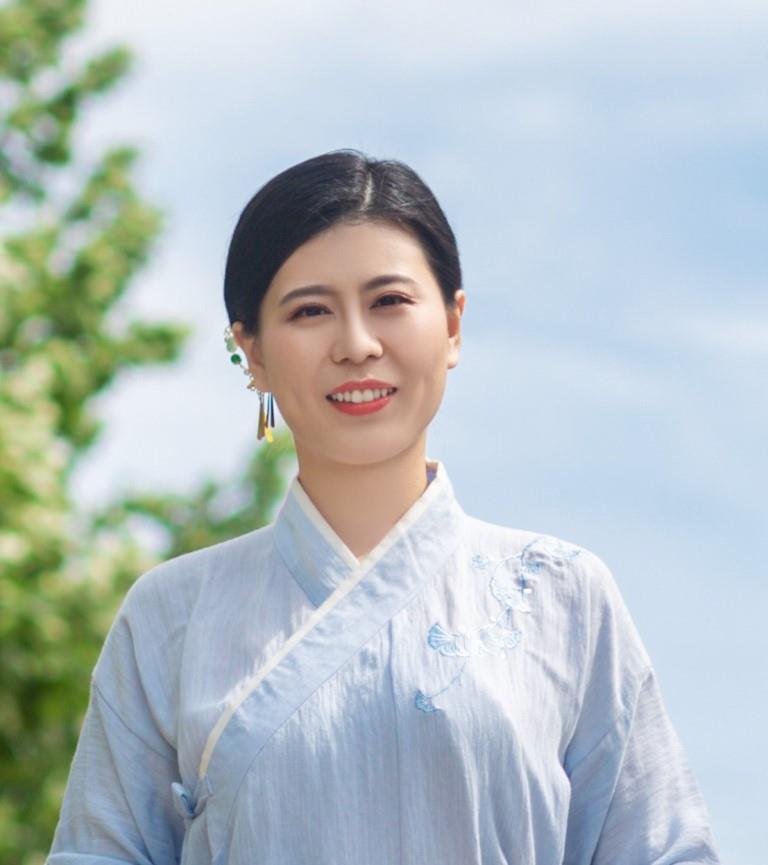}}]{Liyuan Pan} is currently pursuing the Ph.D. degree in the College of Engineering and Computer  Science,  Australian  National  University, Canberra,  Australia.  She received her B.Eng degree from Northwestern Polytechnical University, Xi'an, China in 2014. Her interests include deblurring, flow estimation, depth estimation, and event-based vision. 
\end{IEEEbiography}

\begin{IEEEbiography}[{\includegraphics[width=1in,height=1.25in,clip,keepaspectratio]{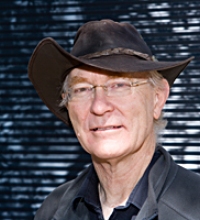}}]{Richard Hartley}
is a member of the computer vision group in the Research School of Engineering, at ANU, where he has been since January, 2001. He is also a member of the computer vision research group in NICTA. He worked at the GE Research and Development Center from 1985 to 2001, working first in VLSI design, and later in computer vision. He became involved with Image Understanding and Scene Reconstruction working with GE’s Simulation and Control Systems Division. He is an author (with A. Zisserman) of the book Multiple View Geometry in Computer Vision. He is a fellow of the IEEE.
\end{IEEEbiography}
\vspace{-15 mm}
\begin{IEEEbiography}[{\includegraphics[width=1in,height=1.25in,clip,keepaspectratio]{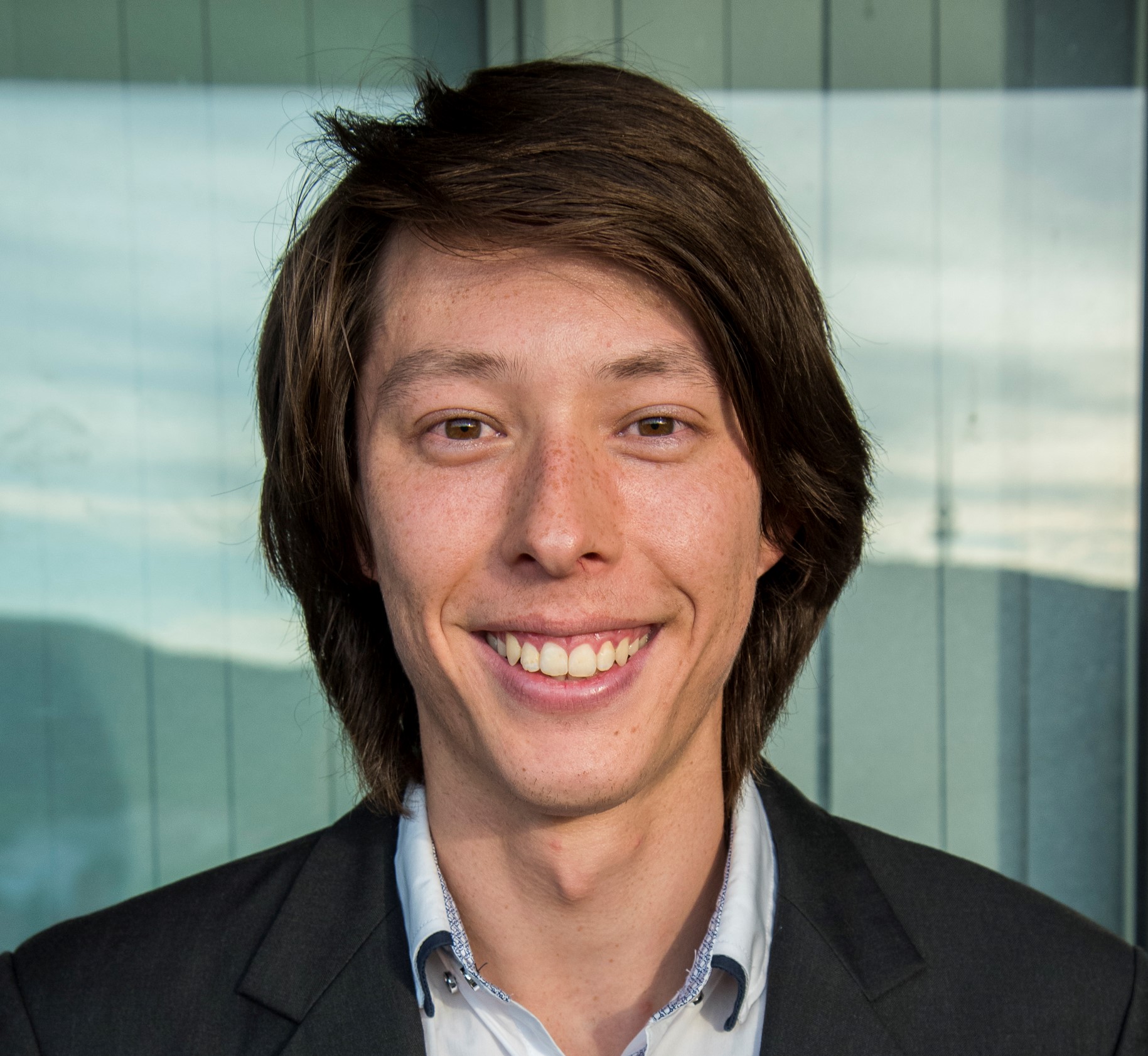}}]{Cedric Scheerlinck}
is currently pursuing the Ph.D. degree in the College of Engineering and Computer  Science,  Australian  National  University, Canberra,  Australia.
He received his BSc and MEng degrees from the University of Melbourne in 2014, 2016. His interests include event-based vision and deep learning.
\end{IEEEbiography}
\vspace{-15 mm}
\begin{IEEEbiography}[{\includegraphics[width=1in,height=1.25in,clip,keepaspectratio]{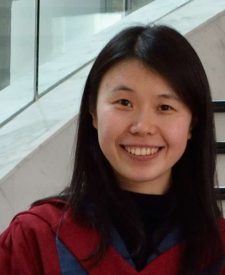}}]{Miaomiao Liu}
received the BEng, MEng, and PhD
degrees from Yantai Normal University, Yantai,
China, Nanjing University of Aeronautics and
Astronautics, Nanjing, China, and the University
of Hong Kong, Hong Kong, China in 2004, 2007, and
2012, respectively. She worked as a researcher in the computer
vision group at
NICTA (2012-2016) and a research scientist (2017-2018) at CSIRO's Data61 in Canberra, Australia. She joined the Australian National University (ANU) as an ARC DECRA fellow in 2018. She is currently also holding a Tenure-track Lecturer position in ANU. She is a member of the IEEE.
\end{IEEEbiography}
\vspace{-15 mm}
\begin{IEEEbiography}[{\includegraphics[width=1in,height=1.25in,clip,keepaspectratio]{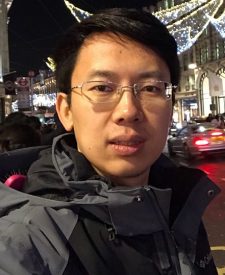}}]{Xin Yu}
received his B.S. degree in Electronic Engineering from University of Electronic Science and Technology of China, Chengdu, China, in 2009, and received his Ph.D. degree in the Department of Electronic Engineering, Tsinghua University, Beijing, China, in 2015. He also received a Ph.D. degree in the College of Engineering and Computer Science, Australian National University, Canberra, Australia, in 2019. He is currently a lecturer in University of Technology Sydney. His interests include computer vision and image processing.
\end{IEEEbiography}
\vspace{-15 mm}
\begin{IEEEbiography}[{\includegraphics[width=1in,height=1.25in,clip,keepaspectratio]{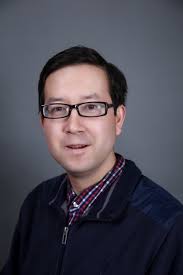}}]{Yuchao Dai} 
is currently a Professor with School of Electronics and Information at the Northwestern Polytechnical University (NPU). He received the B.E. degree, M.E degree and Ph.D. degree all in signal and information processing from Northwestern Polytechnical University, Xi'an, China, in 2005, 2008 and 2012, respectively. He was an ARC DECRA Fellow with the Research School of Engineering at the Australian National University, Canberra, Australia. His research interests include structure from motion, multi-view geometry, low-level computer vision, deep learning, compressive sensing and optimization. He won the Best Paper Award at IEEE CVPR 2012, the Best Paper Award Nominee at IEEE CVPR 2020, the DSTO Best Fundamental Contribution to Image Processing Paper Prize at DICTA 2014, the Best Algorithm Prize in NRSFM Challenge at CVPR 2017, the Best Student Paper Prize at DICTA 2017 and the Best Deep/Machine Learning Paper Prize at APSIPA ASC 2017. He served as Area Chair for IEEE CVPR, ACCV, ACM MM and etc.
\end{IEEEbiography}





\end{document}